\documentclass{article}
\usepackage{graphicx}
\usepackage[draft]{todonotes}
\usepackage{subcaption}
\usepackage{mathtools}
\usepackage{amsfonts}
\usepackage{footnote}
\usepackage{wrapfig}
\usepackage{adjustbox}
\usepackage{multirow}
\usepackage{hyperref}
\usepackage[final]{corl_2020} 
\DeclarePairedDelimiter{\ceil}{\lceil}{\rceil}
\newcommand{\spheading}[2][6em]{
  \rotatebox{90}{\parbox{#1}{\raggedright #2}}}
\newcommand{\printfnsymbol}[1]{%
  \textsuperscript{\@fnsymbol{#1}}%
}
\title{S3CNet: A Sparse Semantic Scene Completion Network for LiDAR Point Clouds}

\author{
Ran Cheng$^{1*}$, Christopher Agia$^{2*}$, Yuan Ren$^{1}$, Xinhai Li$^{1}$, Liu Bingbing$^{1}$
\thanks{$^*$ Indicates equal contribution.}
\thanks{$^1$ are with Huawei, Noah Ark Lab. (e-mail: {ran.cheng1, yuan.ren3, xinhai.li, liu.bingbing}@huawei.com).}
\thanks{$^2$ is with University of Toronto, Toronto. (e-mail: christopher.agia@mail.utoronto.ca).}
}

\begin{document}
\maketitle


\begin{abstract}
    With the increasing reliance of self-driving and similar robotic systems on robust 3D vision, the processing of LiDAR scans with deep convolutional neural networks has become a trend in academia and industry alike. Prior attempts on the challenging Semantic Scene Completion task - which entails the inference of dense 3D structure and associated semantic labels from "sparse" representations - have been, to a degree, successful in small indoor scenes when provided with dense point clouds or dense depth maps often fused with semantic segmentation maps from RGB images. However, the performance of these systems drop drastically when applied to large outdoor scenes characterized by dynamic and exponentially sparser conditions. Likewise, processing of the entire sparse volume becomes infeasible due to memory limitations and workarounds introduce computational inefficiency as practitioners are forced to divide the overall volume into multiple equal segments and infer on each individually, rendering real-time performance impossible. In this work, we formulate a method that subsumes the sparsity of large-scale environments and present S3CNet, a sparse convolution based neural network that predicts the semantically completed scene from a single, unified LiDAR point cloud. We show that our proposed method outperforms all counterparts on the 3D task, achieving state-of-the art results on the SemanticKITTI benchmark \cite{behley2019semantickitti}. Furthermore, we propose a 2D variant of S3CNet with a multi-view fusion strategy to complement our 3D network, providing robustness to occlusions and extreme sparsity in distant regions. We conduct experiments for the 2D semantic scene completion task and compare the results of our sparse 2D network against several leading LiDAR segmentation models adapted for bird's eye view segmentation on two open-source datasets.
\end{abstract}



\section{Introduction}
	
Scene understanding is a challenging component of the autonomous driving problem, and is considered by many as a foundational building block of a complete self-driving system. The construction of maps, the process of locating static and dynamic objects, and the response to the environment during self-driving are inseparable from the agent's understanding of the 3D scene. In practice, scene understanding is typically approached by semantic segmentation. When working with sparse LiDAR scans, scene understanding is not only reflected in the semantic segmentation of the 3D point cloud but also includes the prediction and completion of certain regions, that is, semantic scene completion. Scene completion is the premise of 3D semantic map construction and is a hot topic in current research. However, as semantic segmentation based on 2D image content has reached very mature levels, methods that infer complete structure and semantics from 3D point cloud scenes, which is of significant import to robust perception and autonomous driving, are only at preliminary development and exploration stages. Limited by the sparsity of point cloud data and the lack of features, it becomes very difficult to extract useful semantic information in 3D scenes. Therefore, the understanding of large scale scenes based on 3D point clouds has become an industry endeavor.

\setlength{\columnsep}{5pt}%
\begin{wrapfigure}{r}{8cm}
  \centering
  \subfloat{\includegraphics[trim=0pt 12pt 0pt 0pt,width=\linewidth]{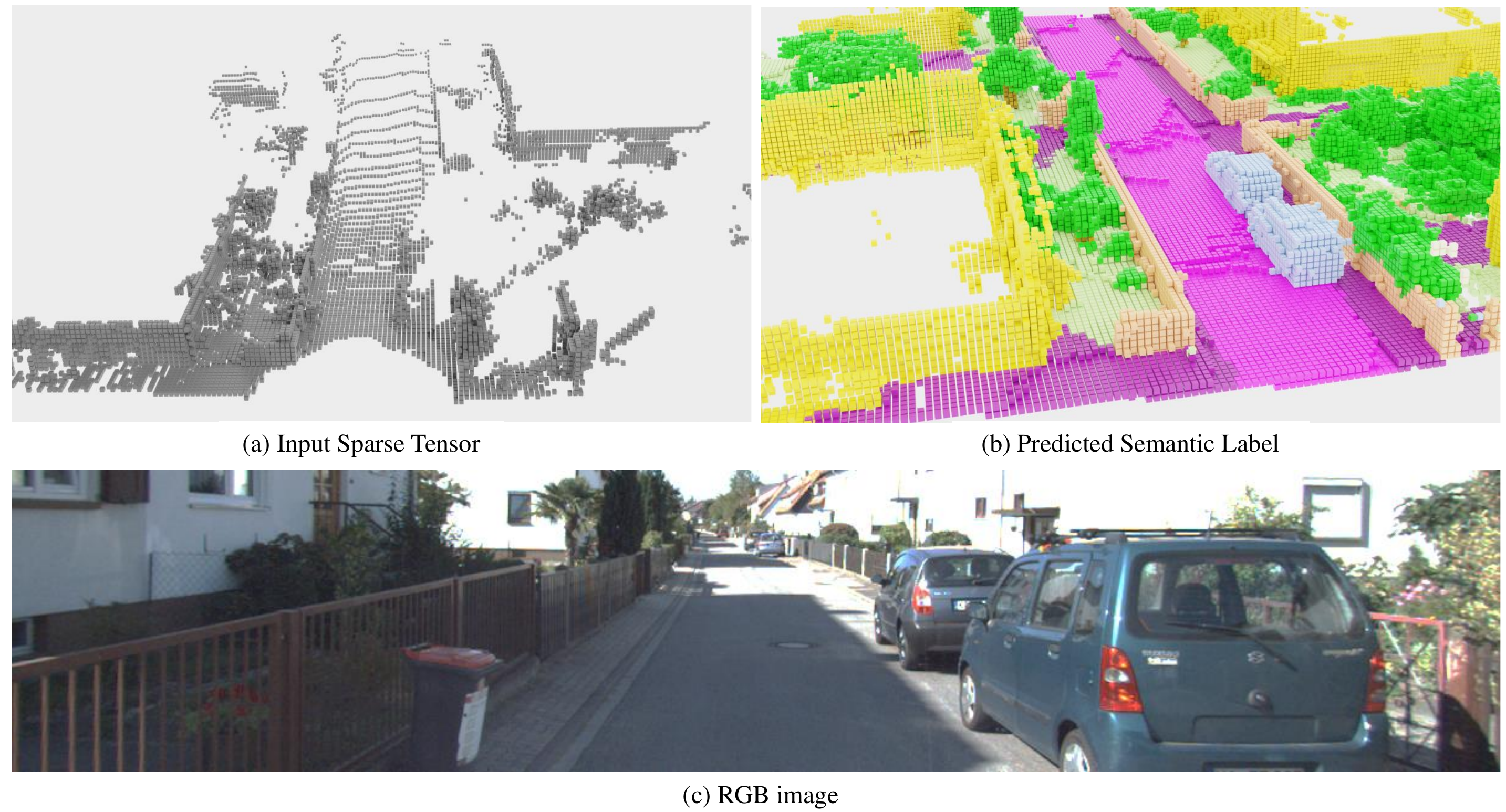}}\\
  \captionsetup{font=scriptsize,labelfont=scriptsize}
  \caption{Semantic Scene Completion on SemanticKITTI Dataset.}
  \label{fig:intro_prediction_gt_result}
  \vspace{-10px}
\end{wrapfigure}


Semantic segmentation and scene completion of 3D point clouds are usually studied separately \cite{gupta2013perceptual, firman2016structured}, but with the emergence of large-scale datasets such as ScanNet \cite{dai2017scannet} and SemanticKITTI \cite{behley2019semantickitti}, researchers have discovered a deep intertwining of an object's semantics with its underlying geometry, and since, have begun exploiting this with the joint learning of semantic segmentation and scene completion to boost model performance \cite{song2017semantic}. For instance, speculating that an object occluded by vehicles and surrounded by leaves is a trunk simplifies the task of inferring it's shape. Conversely, inferring the shape of a pole-like object forms a prior on it's semantic class being a trunk rather than a wall. While previous semantic scene completion methods built on dense 2D or 3D convolutional layers have done well in small-scale indoor environments, they have struggled to maintain their accuracy and efficiency in outdoor environments for several reasons. For one, dense 2D convolutional methods that thrived in the feature rich 2D image space are no longer sufficient when tackling large and sparse LiDAR scans that contain far fewer geometric and semantic descriptors. Furthermore, the dense 3D convolution becomes extremely wasteful in terms of computation and memory since the majority of the 3D volume of interest is in fact empty. Thereby, our main contributions are listed as the following: (a) a sparse tensor based neural network architecture that efficiently learns features from sparse 3D point cloud data and jointly solves the coupled scene completion and semantic segmentation problem; (b) a novel geometric-aware 3D tensor segmentation loss; (c) a multi-view fusion and semantic post-processing strategy addressing the challenges of distant or occluded regions and small-sized objects. Given a single sparse point cloud frame, our model predicts a dense 3D occupancy cuboid with semantic labels assigned to each voxel cell (as shown in Fig. \ref{fig:intro_prediction_gt_result}), generating rich information of the 3D environment that is not contained in the original input such as gaps between LiDAR scans, occluded regions and future scenes.


In order to effectively complete occluded voxel regions from LiDAR scans, we focus on exploiting the geometrical relationship of the 3D points both locally and globally. In this work, we utilize point-wise normal vectors as a geometrical feature encoding to guide our model in filling the gaps according to the object's local surface convexity. We also leverage a LiDAR-based flipped Truncated Signed Distance Function (fTSDF \cite{song2017semantic}) computed from a spherical range image as a spatial encoding to differentiate free, occupied and occluded space of a scene. As for future scenes, because these regions are far from the vehicle and are primarily road or other forms of terrain, we propose a 2D variant of the sparse semantic scene completion network to support the construction of the 3D scene via multi-view fusion with Bird's Eye View (BEV) semantic map predictions. To tackle sparsity, we leveraged the Minkowski Engine \cite{choy20194d}, an auto-differentiation library for sparse tensors to build our 2D and 3D semantic scene completion network. We have also adopted a combined geometric inspired semantic segmentation loss to improve the accuracy of semantic label predictions. Since our network is trained in a complex real-world autonomous driving dataset with 20 classes of dynamic and static objects, and the input data is simply a voxelized LiDAR point cloud appended with geometrical and spatial feature encodings, our model can be deployed on-the-go with various LiDAR sensors. We demonstrate this by applying our method to unseen real-world voxel data, which yields reasonable qualitative results. Our experiments show that our model outperforms all baseline methods by a large margin, with exceptional performance in the prediction of small, under-represented class categories such as bicycles, pedestrians, traffic signs and more.



\section{Related Works}
\label{sec:related_works}
We review the related works across four major areas: volume reconstruction, point cloud segmentation, semantic scene completion, and multi-view fusion.

\textbf{Volume Reconstruction.} There are several approaches to inferring complete volumetric occupancy of shapes and scenes from partial or sparse geometric data. Efficient methods based on object symmetry \cite{kmyg_acquireIndoor_sigga12, schiebener2016heuristic} and plane fitting \cite{monszpart2015rapter} apply for small non-complex completion tasks. In larger scenes with irregular objects, the former are supplanted by methods that fit 3D mesh models to object instances based on the local scene geometry \cite{gupta2013perceptual, gupta2015aligning, geiger2015joint, li2015database}. Yet, a lack of diversity within the 3D model library often leads to incomplete reconstruction, and expanding the library slows down retrieval. Attempts to simplify the process to 3D bounding box fitting neglects the local geometry of objects \cite{jiang2013linear, song2016deep}. Other studies process grid-octree data with CNNs to predict high resolution outputs \cite{riegler2017octnetfusion, tatarchenko2017octree}, but are tailored to reconstructing individual objects rather than entire scenes.

\setlength{\columnsep}{5pt}%
\begin{wrapfigure}{r}{8cm}
  \centering
  \subfloat{\includegraphics[trim=0pt 12pt 0pt 0pt,width=\linewidth]{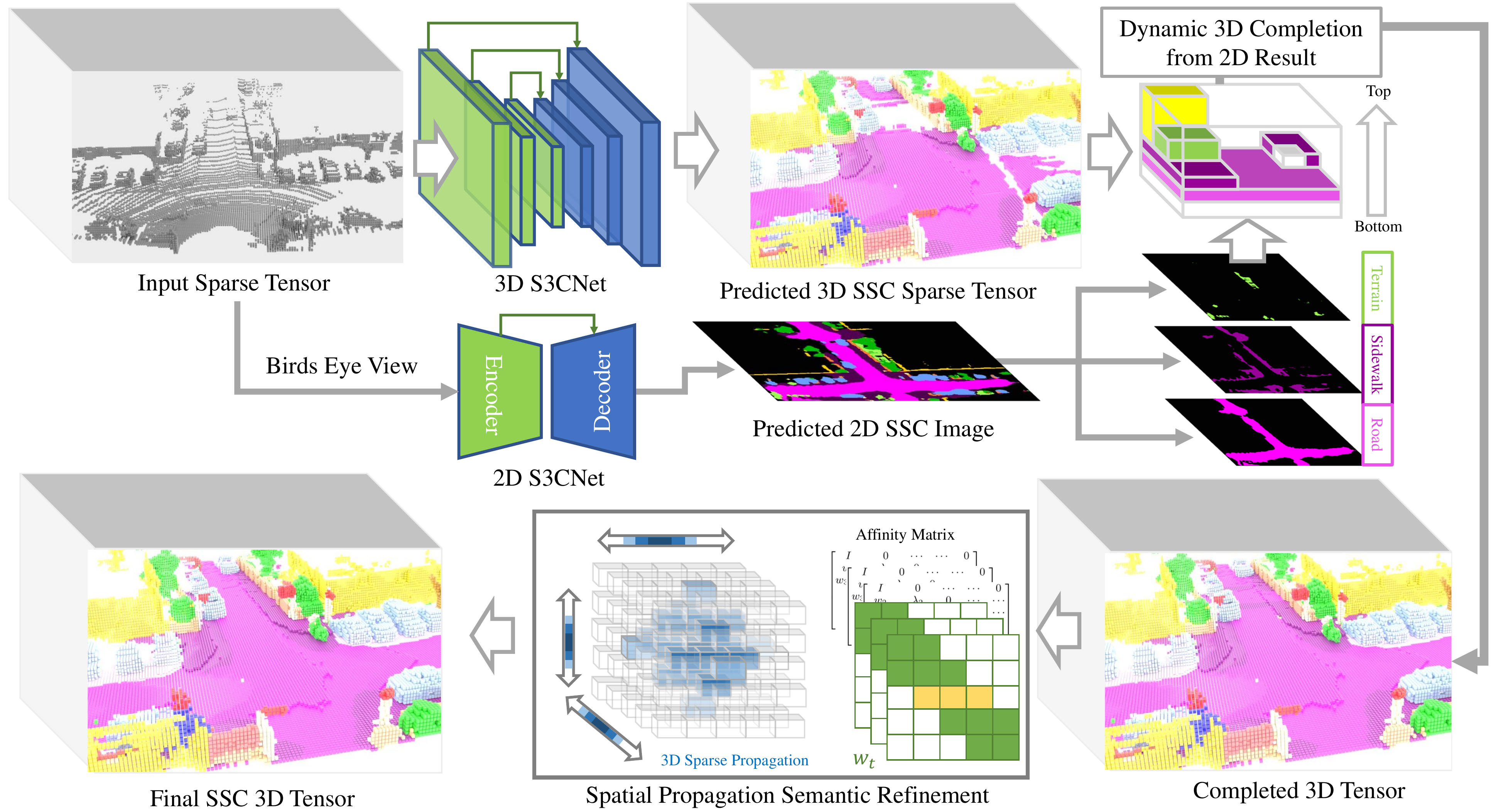}}\\
  \captionsetup{font=scriptsize,labelfont=scriptsize}
  \caption{Full system pipeline.}
  \label{fig:system_pipeline}
  \vspace{-10px}
\end{wrapfigure}

\textbf{Point Cloud Segmentation.} A variety of methods segment range images constructed by spherical projection of a point cloud with deep networks, and back-project the predicted semantic classes onto the corresponding points in 3D space \cite{wu2019squeezesegv2, wang2018pointseg, dewan2019deeptemporalseg}. The small sized range image tensor that these methods operate on leads to unparalleled speed, but $\sim25\%$ of the original point cloud is unrecoverable after the initial spherical projection. Alternative approaches project point clouds onto a bird's eye view perspective, computing pillar features (voxels) from 3D points to construct a BEV map and process it with deep CNNs \cite{yang2018hdnet, aksoy2019salsanet}. Another emergent stream aims to segment point clouds by hierarchically extracting features from the 3D points directly to capture local and global context of the scan \cite{qi2017pointnet++, hu2020randla}. While these methods have had reasonable success, they are typically an order of magnitude slower than their non-direct counterparts. 



\textbf{Semantic Scene Completion.} Most mainstream methods are built on small indoor scenes with dense depth maps or dense point clouds, which effectively reduces the impact of sparseness on scene completion. The representative methods are SSCNet \cite{song2017semantic} and ScanNet \cite{dai2017scannet}. Their approach transforms dense depth maps into a volumetric TSDF signal and passes it through a dense 3D network. Extensions are made in TS3D \cite{garbade2019two} by incorporating semantic segmentation from RGB images into the construction of the input TSDF volume. In outdoor scenes, these methods are limited by the vast increase in sparsity and their network architectures that predict at reduced voxel resolutions. Improvements are made in SATNet \cite{liu2018see}, which maintains high output resolution with a series of dilated convolutional modules (TNet) to capture global context, and propose configurations for early and late fusion of semantic information. \citet{behley2019semantickitti} produced a state-of-the-art system by integrating semantic priors from color images and LiDAR segmentation with the TSDF volume, and adopting a TNet network backbone. However, their memory intensive design requires users to infer on 6 equal parts of the input before fusing the semantic completed partial scenes, deterring potential real-time usage. Furthermore, the dependence on color images introduces instability in the overall system under low-light and poor weather conditions. Therefore, the design of a semantic scene completion method based solely on 3D point cloud data is motivated by the need for faster inference speeds, a smaller memory footprint, and robustness to extreme conditions.

\textbf{Multi-view Fusion.} Fusing semantic and geometric features across various view-points and dimensions has been explored for a variety of tasks. Several of the aforementioned semantic scene completion methods demonstrate the lifting of image-space semantic labels into 3D space \cite{garbade2019two,liu2018see}. Similarly, \citet{hane2013joint} proposed a joint optimization strategy for semantic image segmentation and scene reconstruction, using the completed 3D scene as a geometric prior on the corresponding image pixels to enforce spatially consistent segmentations. Extensions were made to support city-scale reconstruction at reasonable memory costs with an adaptive multi-resolution model \cite{blaha2016large}. \citet{dai20183dmv} designed an 2D-3D network that fuses semantic features from RGB-D images with a differential backprojection layer, achieving multiple view-point reconstruction in indoor settings. Meanwhile, SLAM-based approaches \cite{tateno20162} aim to produce temporally consistent reconstructions by tracking motion states over sequential frames and mapping predicted semantics into 3D space.


\section{Method}
\label{sec:method}

We describe our methods for LiDAR-based semantic scene completion in large outdoor driving scenes. After a brief system overview, we present our unique procedure for computing key spatial features from an input LiDAR scan, the detailed design of our networks, fusion module and refinement module, and a novel loss function incorporating a geometric-aware 3D segmentation loss.

\subsection{System Pipeline}


The entire system pipeline is shown in Fig. \ref{fig:system_pipeline}. From a single LiDAR scan, we construct two sparse tensors that encapsulate the scene into memory efficient 2D and 3D representations. Each tensor is passed through their corresponding semantic scene completion network, 2D S3CNet or 3D S3CNet, to semantically complete the scene in the respective dimension. We propose a dynamic voxel fusion method to further densify the reconstructed scene with the predicted 2D semantic BEV map (detailed discussion in Section \ref{sec:network_arch}). This offsets the significant memory demands on the 3D network - exponential sparsity growth in 3D space makes it difficult to complete classes at range. Using a sparse tensor spatial propagation network \cite{liu2017learning}, we refine the semantic labels in noisy regions of the fused 2D-3D predictions.




\subsection{Spatial Feature Engineering}
\label{sec:feature_engineering}

\setlength{\columnsep}{1pt}%
\begin{wrapfigure}{r}{7cm}
  \centering
  \includegraphics[trim=0pt 5pt 0pt 0pt,width=\linewidth]{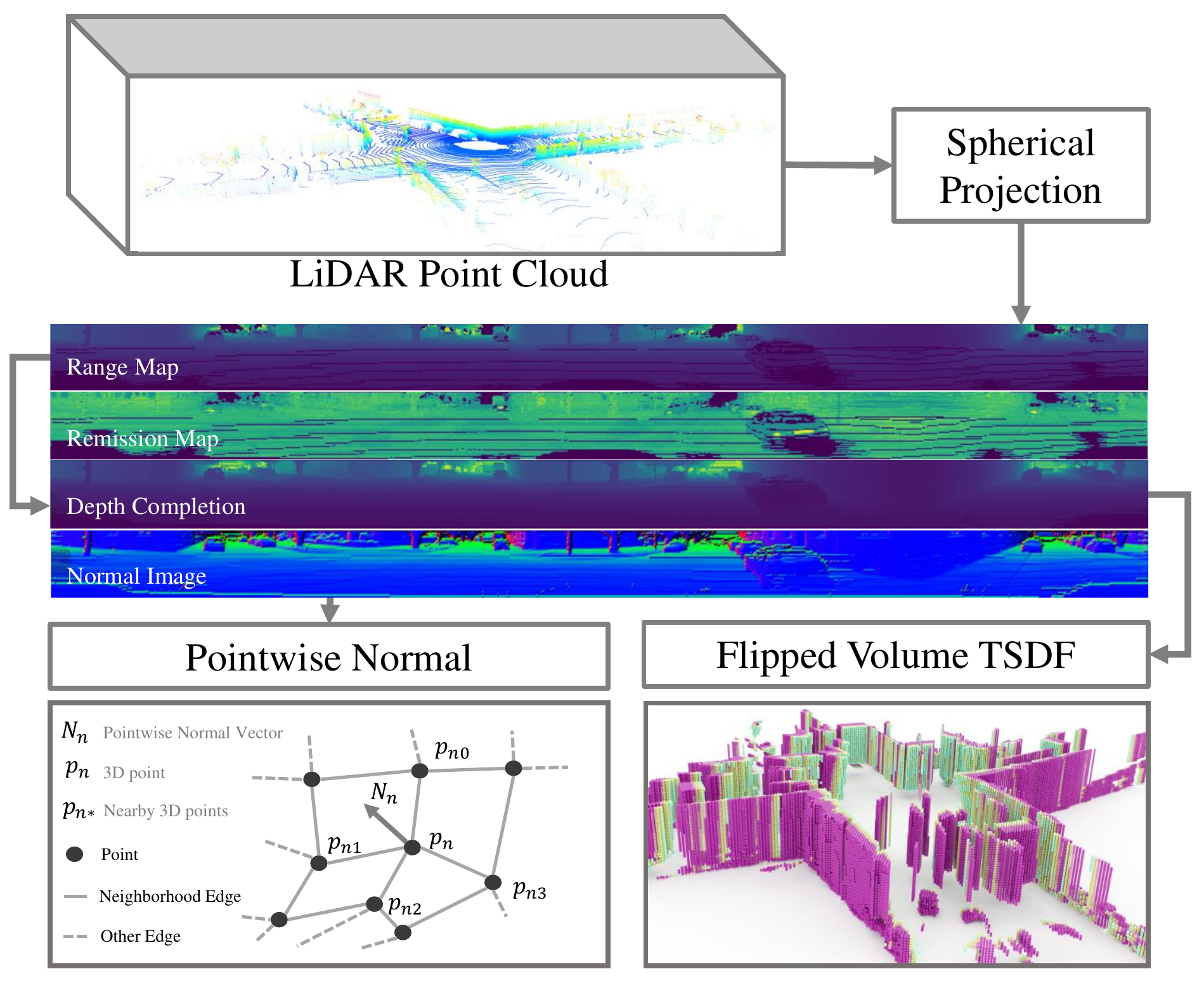}
  \captionsetup{font=scriptsize,labelfont=scriptsize}
  \caption{Sparse 3D tensor features.}
  \label{fig:sparse_tensor_features}
  \vspace{-10px}
\end{wrapfigure}

\textbf{Sparse 2D Feature.} We define the sparse 2D tensor as the set of non-empty pillars approximating the point distribution along the $x$-$y$ plane (BEV). Each pillar, $p_{i,j}$, is a 7-dimensional vector encoding the mean, minimum, and maximum heights and intensities of points within the voxel, and the point density; all values are normalized. 


\textbf{Sparse 3D Feature.} The sparsity of the raw point cloud makes it difficult to extract spatial features, and thus, we transform it into a range image via spherical projection. As the quality of extracted spatial features are sensitive to noise, we retrieve a smooth range image by performing dynamic depth completion using the dilation method of \citet{ku2018defense}. This enables robust extraction of a 3-dimensional normal surface feature for each pixel that is reversely assigned to the points in 3D space. Further, we maintain a memory efficient sparse 3D tensor by modifying the sign-flipped TSDF approach of \citet{song2017semantic}, and compute TSDF values from the smooth range image, storing only the coordinates within the truncation range of existing LiDAR points. All other features for TSDF generated coordinates are zero-padded.

\subsection{Network Architecture}
\label{sec:network_arch}

We adopt the Minkowski Engine \cite{choy20194d} as our sparse tensor auto-differentiation framework to build the entire system. A sparse tensor can be defined as a hash-table of coordinates and their corresponding features: $\mathbf{x} = [\mathbf{C}_{n \times d}, \mathbf{F}_{n \times m}]$. Here, $n$ are the number of non-empty voxels, $d$ and $m$ are the dimension of coordinates and features, respectively. The sparse convolutional layer is thus:

\begin{equation}
    \mathbf{x}_{\mathbf{u}} = \sum_{\mathbf{i} \in \mathcal{N}^D(u\mathbf{u}, K, \mathcal{C}^{in})}{\mathbf{W_i}\mathbf{x_{\mathbf{u+i}}}} \textrm{ for } \mathbf{u} \in \mathcal{C}^{out} 
\end{equation}

Where $K$ is the kernel size and $\mathcal{N}^D(\mathbf{u}, K, \mathcal{C}^{in})$ are the set of offsets that are at most $\ceil{\frac{1}{2}(K - 1)}$ away from $\mathbf{u}$, the current coordinate. Unlike the conventional convolution, this generalized convolution (introduced by \citet{choy20194d} \emph{et al}) suits generic input and output coordinates, and arbitrary kernel shapes. It allows extending a sparse tensor network to extremely high-dimensional spaces and dynamically generating coordinates for generative tasks. The arbitrary input shape and multi-dimensional support enables us deploy the same network layout in different dimensional spaces. Hence, our 2D and 3D networks share the same set of network components (built off sparse convolution, transposed convolution and pooling layers) with differing coordinate dimensions.

\setlength{\columnsep}{5pt}%
\begin{wrapfigure}{l}{7cm}
  \centering
  \captionsetup{font=scriptsize,labelfont=scriptsize}
  \subfloat{\includegraphics[trim=0pt 12pt 0pt 0pt,width=\linewidth]{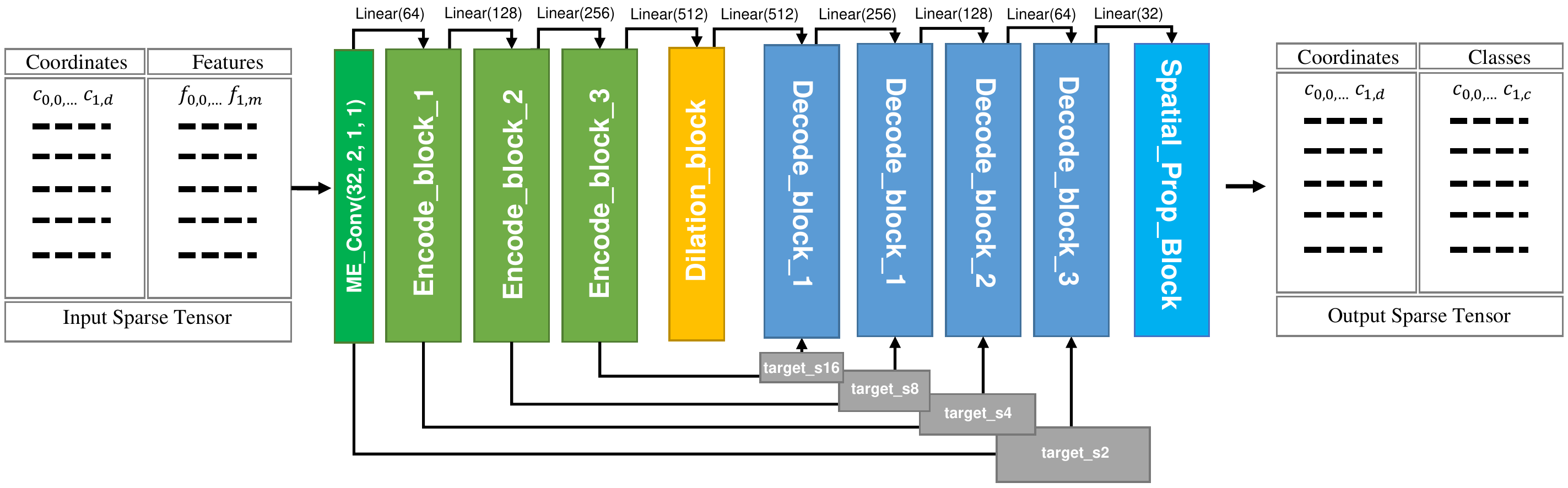}}\\
  \subfloat{\includegraphics[trim=0pt 12pt 0pt 0pt,width=0.45\linewidth]{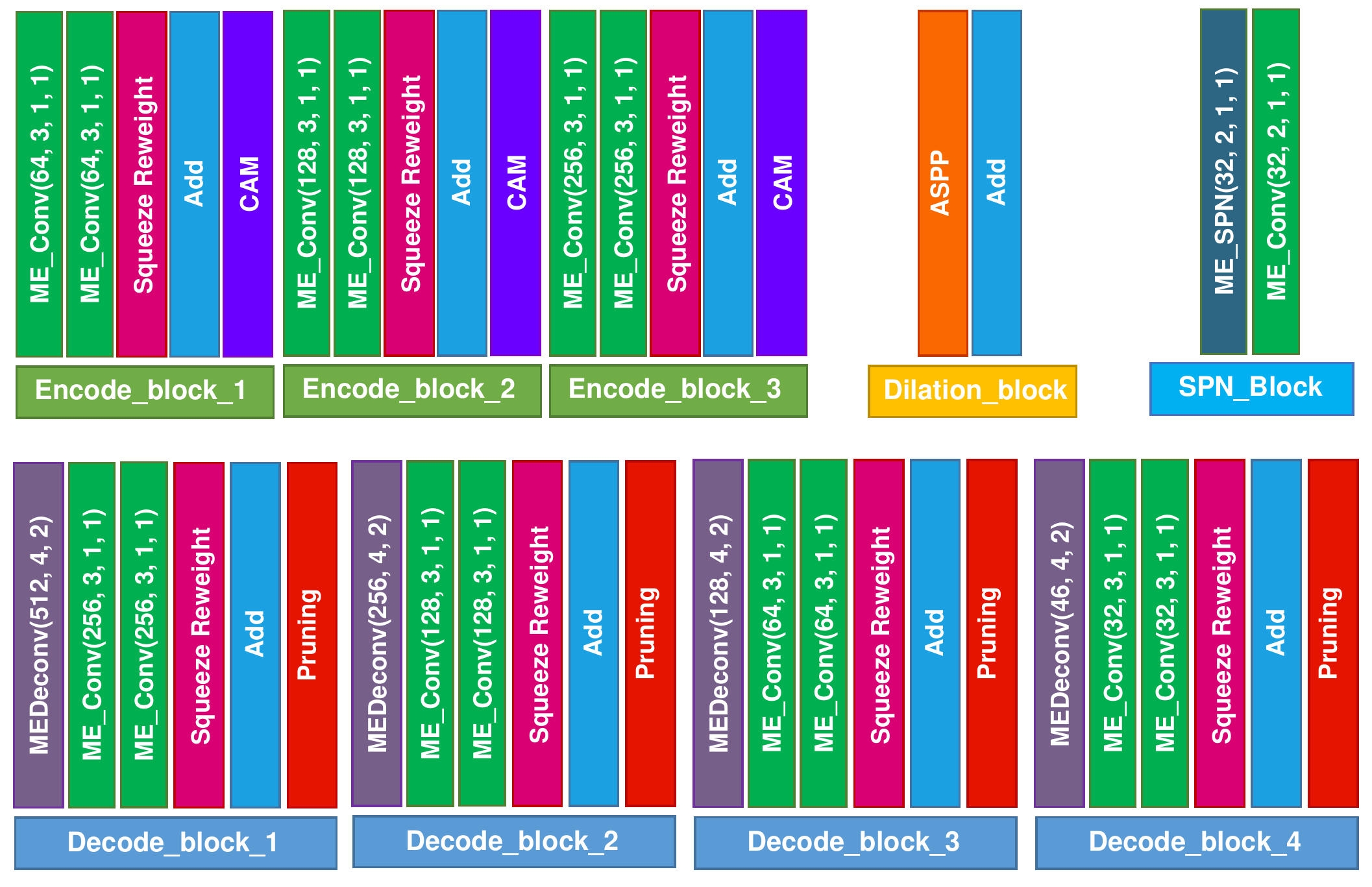}}
  \subfloat{\includegraphics[trim=0pt 12pt 0pt 0pt,width=0.45\linewidth]{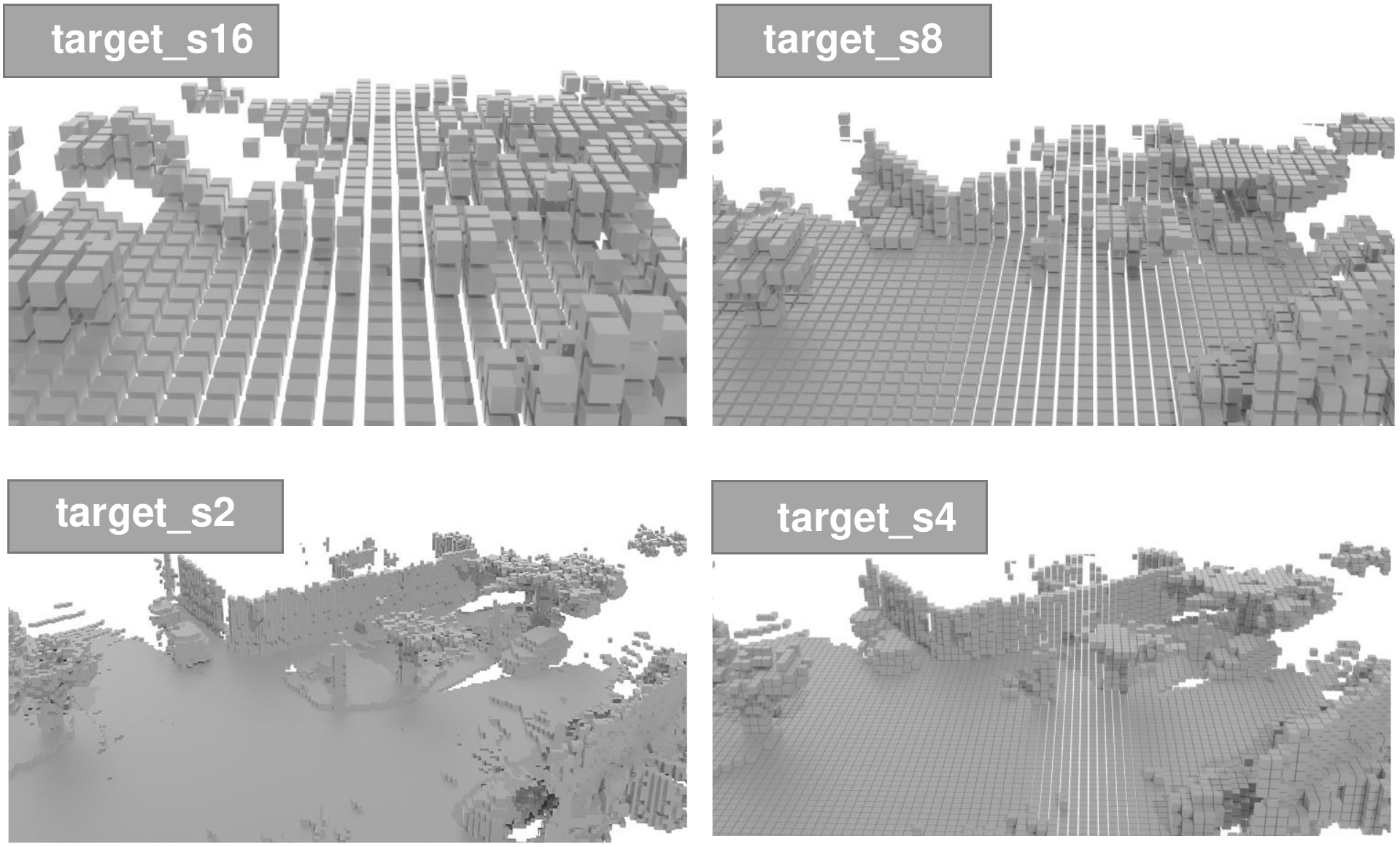}}
  \captionsetup{font=scriptsize,labelfont=scriptsize}
  \caption{S3CNet network architecture. Example target masks are visualized as target$\_$s2-16.}
  \label{fig:system_architecture}
  \vspace{-10px}
\end{wrapfigure}


After extracting the features as described in Section \ref{sec:feature_engineering}, we create the sparse 3D tensor by voxelizing the point cloud and extracting the coordinates and features of non-empty or TSDF generated points. If multiple points occupy the same voxel, their features are averaged. The voxel resolution is 0.2m in each dimension; this applies to the sparse 2D tensor as well. The spatial extent of our predictions in the $x$-$y$-$z$ directions are [0m, 51.2m], [-25.6m, 25.6m], [-2m, 4.4m], respectively. Discretizing the volume yields a [256, 256, 32] sized tensor, upon which our sparse 3D network will predict the set of occupied coordinates and a probability distribution over the 20 possible semantic categories.

\setlength{\columnsep}{5pt}%
\begin{wrapfigure}{r}{5cm}
  \centering
  \captionsetup{font=scriptsize,labelfont=scriptsize}
  \subfloat[Local geometric anisotropy of a sparse tensor voxel $c_p$ in the predicted 3D semantic scene completion label.]{\includegraphics[trim=0pt 12pt 0pt 0pt,width=\linewidth]{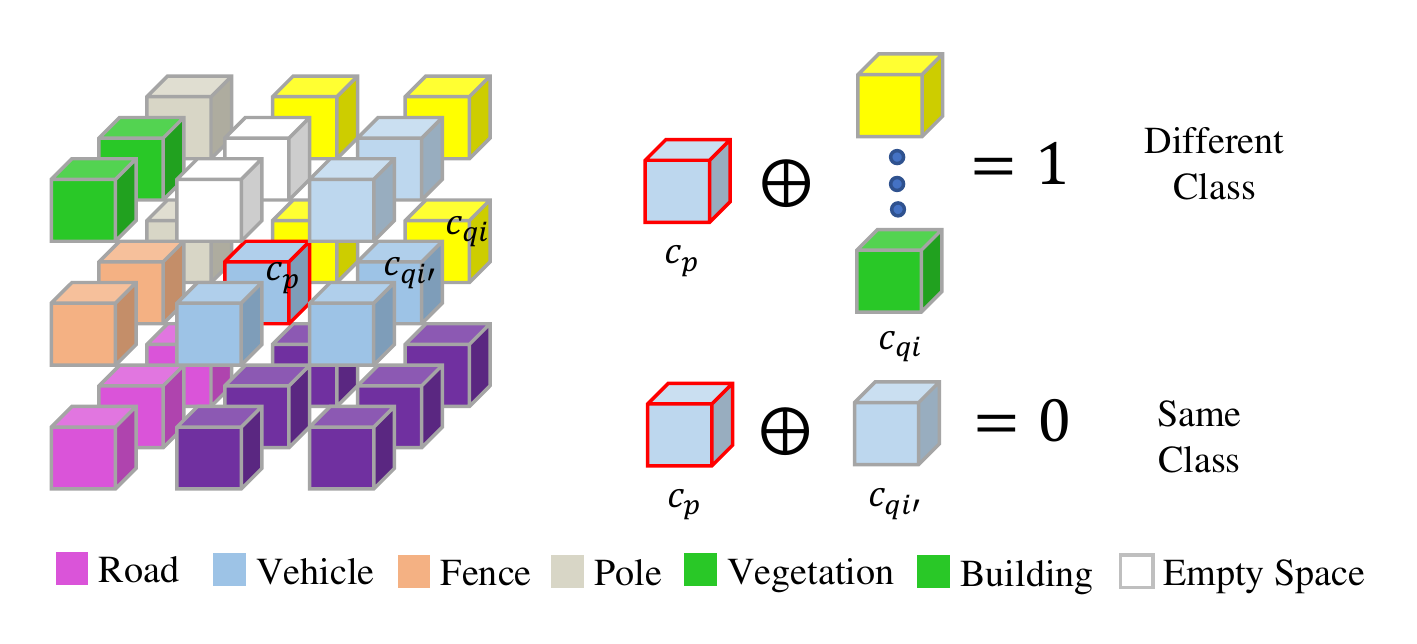}}\\
  \subfloat[Gradient of tensor based on anistropy operation.]{\includegraphics[trim=0pt 12pt 0pt 0pt,width=\linewidth]{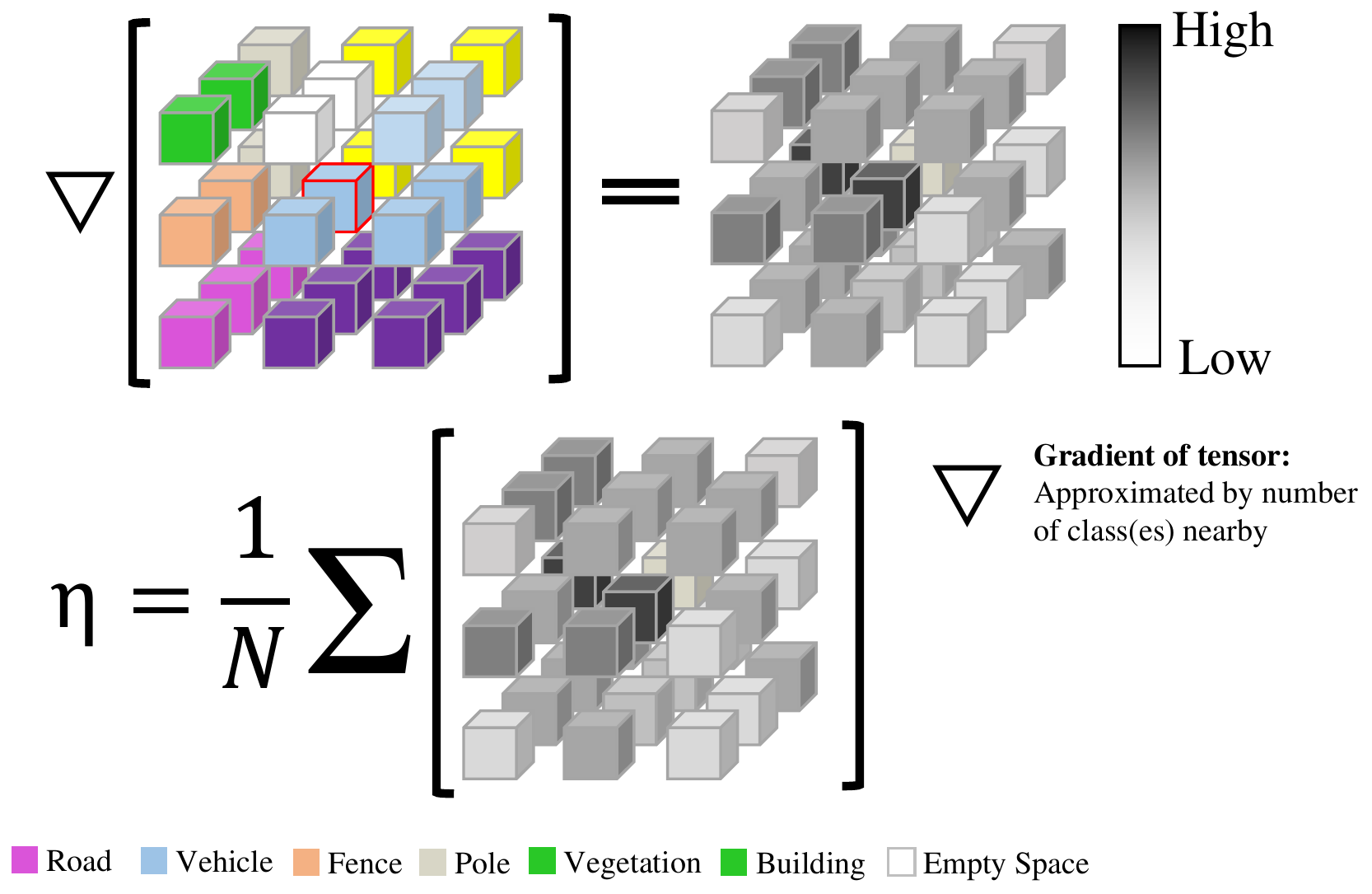}}
  \captionsetup{font=scriptsize,labelfont=scriptsize}
  \caption{Diagrams that illustrate how to calculate the local geometric anisotropy and gradient for a given prediction label voxel grid.}
  \label{fig: lga_vis}
  \vspace{-10px}
\end{wrapfigure}


As illustrated in Fig. \ref{fig:system_architecture}, our 3D network is assembled from four primary building blocks: encode, decode, dilation, and spatial propagation. We implement a squeeze re-weight (SR) \cite{wang2018pointseg} layer to model inter-channel dependencies and improve generalization. Each encoder block contains a context aggregation module (CAM) \cite{wu2019squeezesegv2} which captures context in a large receptive field, improving robustness to dropout noise. We adapt these modules for sparse tensor support, and observe improvements to both the segmentation and completion tasks. The decode blocks utilize sparse transposed convolutions capable of generating new coordinates with the outer product of the weight kernel and the input coordinates. As new coordinates in 3D are generated by the cube of the kernel size, we preserve memory with pruning modules that remove redundant coordinates throughout scene completion - training supervision is provided by ground truth filter masks (Fig. \ref{fig:system_architecture}, target$\_$s2-16). The dilation block features a sparse atrous spatial pyramid pooling (ASPP) module to trade-off accurate localization (small field-of-view) with context assimilation (large field-of-view), our dilation rates are [2,3,4]. The spatial propagation block contains two parts, 3D spatial propagation module and guidance convolution network. The guidance network produces affinity matrices used to guide the spatial propagation network to deform the sparse tensor into a desired 3D shape.



\textbf{Loss function.} When training our 2D network, we accomplish healthy BEV completion and balanced learning of under-represented class categories by combining pixel-wise focal loss, weighted cross entropy loss. Per-voxel binary cross entropy loss is used to train the pruning modules.

\begin{equation}
    \mathcal{L}_{2D}(p, y) = - (\alpha \sum_{c \in C}^{C}w_c y_c log(p_c) + \beta \sum_{c \in C}^{C} y_c (1 - p_c)^\gamma log(p_c)) + \omega \mathcal{L}_{\mathbf{BCE}}(p, y)
\end{equation}

Here, $p$ is the predicted label and $y$ is the ground truth label. The weighting factors $\alpha$, $\beta$, and $\omega$ are empirically set to 0.5, 0.5, and 1, respectively. We define a novel loss function to train our 3D network. It consists of a completion term (voxelized binary cross entropy) and a geometric-aware 3D tensor segmentation loss. The two terms are balanced by an $\lambda$ constant empirically set to 0.35.

\begin{equation}
    \mathcal{L}_{3D}(p, y) = \lambda \mathcal{L}_{\mathbf{completion}}(p, y) + (1- \lambda) \mathcal{L}_{\mathbf{GA}}(p, y)
\end{equation}

The completion loss is expressed below, where $\mathcal{L}_{\mathbf{BCE}}$ is the binary cross entropy loss over volumetric occupancy. Note that this is applied to train the pruning modules at various scale spaces.

\begin{equation}
    \mathcal{L}_{\mathbf{completion}}(p, y) = \sum_{i,j,k}{\mathcal{L}_{\mathbf{BCE}}(p_{ijk}, y_{ijk})}
\end{equation}

Below is the geometric-aware 3D tensor segmentation loss computed over the final output tensor.

\begin{equation}
    \mathcal{L}_{GA}(p, y) = -\frac{1}{N}\sum_{i,j,k}{\sum_{c=1}^{C}{(\xi + \eta M_{LGA})y_{ijk, c}log(p_{ijk, c})}}
    \label{eq:lga}
\end{equation}

\setlength{\columnsep}{5pt}%
\begin{wrapfigure}{r}{4cm}
  \centering
  \includegraphics[trim=0pt 12pt 0pt 0pt,width=\linewidth]{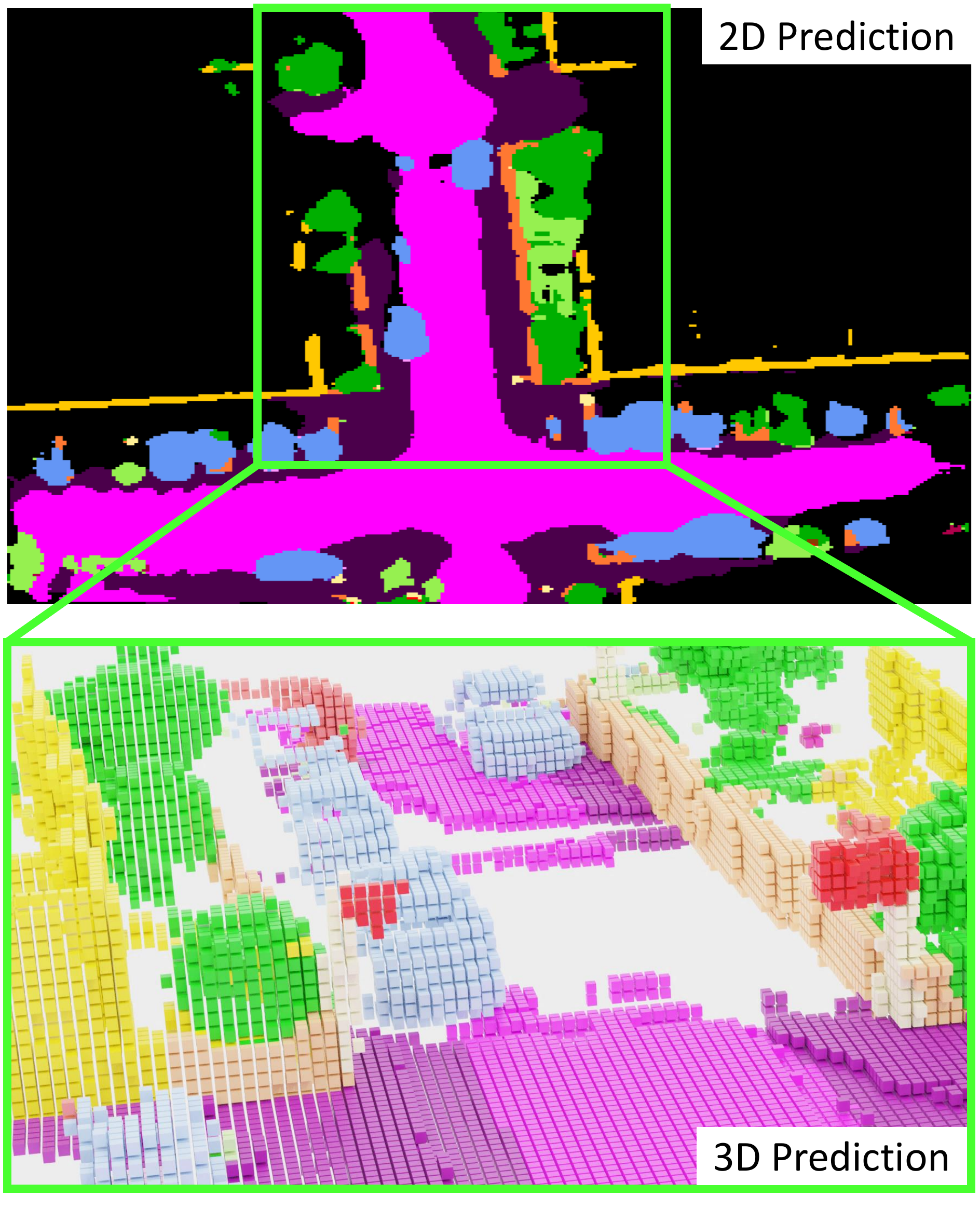}
\captionsetup{font=scriptsize,labelfont=scriptsize}
  \caption{Birds eye view 2D results compared with 3D predictions.}
  \label{fig: 2dvs3d}
  \vspace{-15px}
\end{wrapfigure}

The terms $M_{LGA}$, $\xi$ and $\eta$ describe signals computed from the same local cubic neighborhood of a given coordinate (As shown in Fig. \ref{fig: lga_vis}). The \textit{Local Geometric Anisotropy} defined by \citet{li2019depth}, $M_{LGA} = \sum_{i=1}^{K}{(c_p \oplus c_{qi})}$, is a discrete smoothness promoting signal that penalizes the prediction of many classes within a local neighborhood. Although it ensures locally consistent predictions in homogeneous regions (i.e. road center, middle of a wall), it may incorrectly divert the model from inferring boundaries between separate objects. We thus introduce $\eta$, which accounts for the local arrangement of classes based on the volumetric gradient, and downscales $M_{LGA}$ when the neighborhood contains structured predictions of different classes. To smoothen-out the loss manifold, we include a continuous entropy signal $\xi = -\sum_{c \in C'}{P(c)log(P(c))}$, where $P(c)$ is the distribution of class $c$ amongst all classes $C'$ in the local neighborhood. 

Eq. (\ref{eq:lga}) thus decomposes into two multiplicative factors with the cross-entropy loss which explicitly models the relationship between a predicted voxel and it's local neighborhood. Intuitively, the smooth local entropy term, $\xi$, down-scales the loss in easily classified homogeneous regions, enabling the network to attend to non-uniform regions (e.g. class boundaries) as learning progresses. However, a measure of non-uniformity alone is insufficient in that non-uniform regions should be more heavily penalized if the predicted neighborhood lacks structure. This motivates the inclusion of $\eta M_{LGA}$ which also considers spacial arrangement of classes and down-scales the loss in structured cubic regions. In combination, we acquire a smooth loss manifold when the local prediction is close to the ground truth as well as uncluttered, with sharp increases when the local cubic neighbourhood is noisy and far off from the ground truth. This speeds up convergence while reducing the chance of stabilizing in local optimums.




\textbf{Multi-view Fusion and Spatial Propagation.} 

The task of 2D semantic scene completion is far less complex than it's 3D counterpart because it does not account for the local structure of objects along the $z$ dimension, relegating the task to semantic filling in the bird's eye view plane. This allows our sparse 2D network to more easily understand the global distribution pattern of semantic classes even at far distances, yielding accurate predictions of roads, terrain, and side walks. When confronted with heavy noise or occlusions, such as Fig. \ref{fig: 2dvs3d}, our 3D model may lack the confidence to complete certain regions. To remedy this, we express the predicted 3D tensor as a stack of BEV images, and attempt to fill in each empty voxel with the corresponding prediction in the 2D network. The lifting algorithm operates as follows: (1) split the 3D volume into layers along the z-dimension (32 layers in this application); (2) locate the slice with the maximum occurrence of the desired class; (3) fill each voxel with the associated 2D pixel prediction if there exists such a class in the voxel's $n\times n$ ($n=3$) neighborhood, otherwise attempt to fill in the above voxel until the highest layer is exceeded; (4) repeat for all classes. Hence, by lifting the 2D voxel-wise semantic completion into 3D space, we improve both completion and segmentation metrics of the 3D task at minimal cost. To mitigate any additional post-fusion noise, we apply a spatial propagation network that refines the segmentation results in 3D space.


\section{Experiments}
\label{sec:result}

\begin{figure}[htb]
\vspace{-10px}
    \centering
    \includegraphics[width=\linewidth]{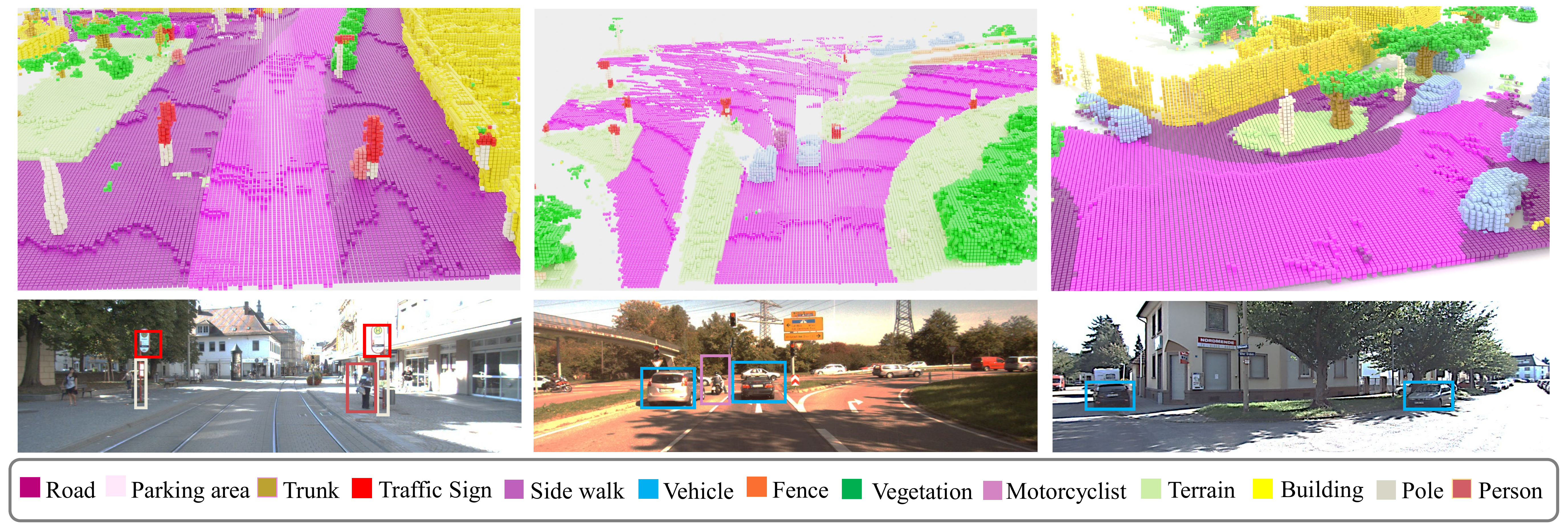}
    \captionsetup{font=scriptsize,labelfont=scriptsize}
    \caption{Qualitative results of predictions on SemanticKITTI test sequences. Best viewed in color.}
    \label{fig:qualitative_result}
\vspace{-10px}
\end{figure}


We evaluate our system on the SemanticKITTI dataset \cite{behley2019semantickitti}, which contains 8,728 voxelized LiDAR scans and the corresponding ground truth labels. We use the public train/val/test split defined in the SemanticKITTI API, and provide qualitative visualizations of our predictions. For evaluation metrics, we follow  \cite{song2017semantic} using mean IoU over positive classes and binary per-voxel completion IoU. Experiments are deployed on Nvidia GP100 GPUs (16GB Graphic RAM); we use two GPUs per model and train for 50 epochs each. The average inference time for 3D S3CNet is 0.5s (frame with 40,000$\pm$500 points), and training requires around 120 hours. For 2D S3CNet, the average inference time is 0.05s resulting in 50 training hours. The training scheme for the best performing 2D and 3D models are: (3D) Adam optimizer, 0.0025 learning rate, 0.0005 weight decay, and (2D) SGD optimizer, 0.001 learning rate, and 0.0005 weight decay. Both experiments also incorporate an exponential learning rate scheduler with a decay rate of 0.9 every 10 epochs.

\setlength{\columnsep}{5pt}%
\begin{wrapfigure}{r}{6cm}
  \centering
  \includegraphics[trim=0pt 12pt 0pt 0pt,width=\linewidth]{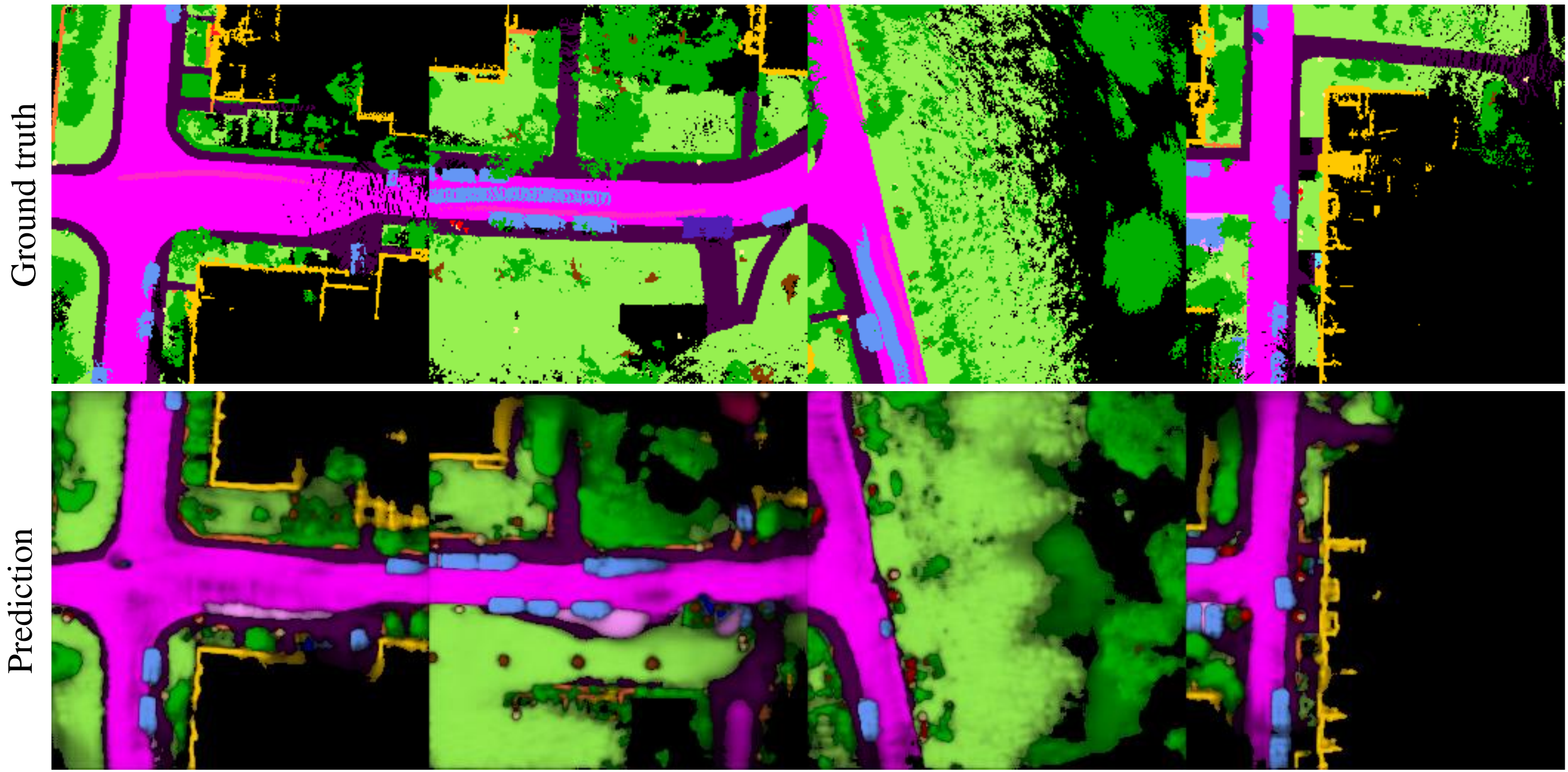}
\captionsetup{font=scriptsize,labelfont=scriptsize}
  \caption{2D qualitative results.}
  \label{fig: 2d_qualitative}
  \vspace{-10px}
\end{wrapfigure}

\textbf{Qualitative Results.} Fig. \ref{fig:qualitative_result} depicts the inferences of our system on three SemanticKITTI test set samples and the corresponding RGB images. We notice that our method correctly captures most classes with exceptional detail and consistency, including small object categories like people and traffic signs in challenging scenes. The middle column highlights an open intersection scene with heavy front-facing occlusion, complicating the task of inferring distant mid-intersection vehicles. For 2D S3CNet, we demonstrate the prediction and ground truth BEV map in static scenes and dynamic scenes with moving vehicles (as shown in Fig. \ref{fig: 2d_qualitative}).

\begin{table}[htb]
\begin{adjustbox}{max width=\textwidth}
\begin{tabular}{|c|c|cccccccccccccccccccc|}
\hline
\multicolumn{1}{|c|}{Approach} & \spheading{\textbf{mIoU}} & \spheading{completion} & \spheading{road} & \spheading{sidewalk} & \spheading{parking} & \spheading{other-ground} & \spheading{building} & \spheading{car} & \spheading{truck} & \spheading{bicycle} & \spheading{motorcycle} & \spheading{other-vehicle} & \spheading{vegetation} & \spheading{trunk} & \spheading{terrain} & \spheading{person} & \spheading{bicyclist} & \spheading{motorcyclist} & \spheading{fence} & \spheading{pole} & \spheading{traffic-sign} \\ \hline
\multicolumn{1}{|c|}{Ours}      & \textbf{0.295 (1)}                 & 0.45 (5)                       & 0.42 (6)                 & 0.22 (6)                     & 0.17 (6)                    & 0.07 (4)                         & \textbf{0.52 (1)}                     & 0.31 (3)                & 0.06 (2)                  & \textbf{0.415 (1)}                    & \textbf{0.45 (1)}                       & \textbf{0.16 (1)}                          & 0.39 (5)                       & \textbf{0.34 (1)}                  & 0.21 (6)                    & \textbf{0.45 (1)}                   & \textbf{0.35 (1)}                      & \textbf{0.16 (1)}                         & \textbf{0.31 (1) }                 & \textbf{0.31 (1)}                 & \textbf{0.24 (1) }                        \\
\multicolumn{1}{|c|}{JS3C-Net} & 0.238 (2)                 & \textbf{0.56 (1)}                       & \textbf{0.64 (1)}                 & \textbf{0.39 (1)}                     & \textbf{0.34 (1) }                   & \textbf{0.14 (1) }                        & 0.394 (2)                     & 0.33 (1)                & \textbf{0.072 (1) }                 & 0.14 (2)                    & 0.08 (2)                       & 0.12 (2)                          & 0.43 (1)                       & 0.19 (5)                  & \textbf{0.40 (1)}                    & 0.08 (2)                   & 0.05 (2)                      & 0.00 (2)                         & 0.30 (2)                  & 0.18 (3)                 & 0.15 (3)                         \\
\multicolumn{1}{|c|}{gjt}      & 0.21 (3)                 & 0.54 (2)                       & 0.61 (4)                 & 0.36 (2)                     & 0.32 (2)                    & 0.06 (5)                         & 0.38 (3)                     & 0.32 (2)                & 0.05 (3)                  & 0.02 (3)                    & 0.03 (4)                       & 0.07 (4)                          & 0.39 (4)                       & 0.18 (6)                  & 0.34 (3)                    & 0.04 (3)                   & 0.01 (4)                      & 0.00 (3)                         & 0.27 (3)                  & 0.15 (6)                 & 0.12 (4)                         \\
\multicolumn{1}{|c|}{ifelse}    & 0.183 (4)                 & 0.51 (3)                       & 0.63 (2)                 & 0.31 (3)                     & 0.27 (3)                    & 0.14 (2)                         & 0.29 (6)                     & 0.29 (5)                & 0.039 (6)                  & 0.00 (5)                    & 0.00 (5)                       & 0.00 (5)                          & 0.40 (2)                       & 0.21 (3)                  & 0.34 (2)                    & 0.00 (5)                   & 0.00 (5)                      & 0.00 (5)                         & 0.25 (4)                  & 0.19 (2)                 & 0.06 (5)                         \\
\multicolumn{1}{|c|}{jbehley \cite{behley2019semantickitti}}  & 0.177 (5)                 & 0.50 (4)                       & 0.62 (3)                 & 0.31 (4)                     & 0.23 (5)                    & 0.06 (6)                         & 0.34 (4)                     & 0.30 (4)                & 0.04 (5)                  & 0.00 (5)                    & 0.00 (5)                       & 0.00 (6)                          & 0.40 (3)                       & 0.21 (2)                  & 0.33 (4)                    & 0.00 (5)                   & 0.00 (5)                      & 0.00 (5)                         & 0.24 (5)                  & 0.16 (4)                 & 0.06 (6)                         \\
\multicolumn{1}{|c|}{yanx27}   & 0.17 (6)                 & 0.41 (6)                       & 0.43 (5)                 & 0.28 (5)                     & 0.26 (4)                    & 0.10 (3)                         & 0.29 (5)                     & 0.26 (6)                & 0.05 (4)                  & 0.00 (4)                    & 0.04 (3)                       & 0.09 (3)                          & 0.35 (6)                       & 0.20 (4)                  & 0.28 (5)                    & 0.02 (4)                   & 0.07 (3)                      & 0.00 (4)                         & 0.23 (6)                  & 0.16 (5)                 & 0.16 (2)                        \\                    
\hline
\multicolumn{15}{l}{\textbf{Source}: https://competitions.codalab.org/competitions/22037}
\end{tabular}
\end{adjustbox}
\captionsetup{font=scriptsize,labelfont=scriptsize}
\caption{SemanticKITTI Test Set Benchmark. ($*$) is the prediction IoU rank.}
\label{table:test_set_benchmark}
\vspace{-10px}
\end{table}

\begin{wraptable}{r}{5cm}
\centering
\begin{adjustbox}{max width=0.35\textwidth}
\begin{tabular}{|c|c|c|}
\hline
\multirow{2}{*}{Model}                     & \multicolumn{2}{c|}{mIoU}     \\  \cline{2-3} 
                                           & SK               & NS               \\ \hline
HDUNet \cite{yang2018hdnet}                & 0.1486           & 0.2777           \\
PointSeg \cite{wang2018pointseg}           & 0.1204           & 0.2317           \\
SSv2 (w/o CRF) \cite{wu2019squeezesegv2}   & 0.1294           & 0.1718           \\
DBLiDARNet \cite{dewan2019deeptemporalseg} & 0.1682           & 0.2382           \\
SalsaNext \cite{cortinhal2020salsanext}    & 0.1780           & 0.2715           \\ \hline
Ours                                       & \textbf{0.27003} & \textbf{ 0.3032} \\ \hline
\end{tabular}
\end{adjustbox}
\captionsetup{font=scriptsize,labelfont=scriptsize}
\caption{2D S3CNet quantitative results and comparison to state-of-the-art on SemanticKITTI validation set (SK) and nuScenes test set (NS). \cite{caesar2020nuscenes})}
\label{table:2d_quantitative_result}
\vspace{-10px}
\end{wraptable}

\textbf{Quantitative results.} Our primary results are shown in Table. \ref{table:test_set_benchmark}, where we compare our overall system to several state-of-the-art methods in the SemanticKITTI test set benchmark. Note the approaches without citations are non-published works. At the time of writing our proposed S3CNet considerably outperforms all others, achieving a mean IoU score of 29.5$\%$ (a +23.9$\%$ improvement on the previous leading method) and a +66.6$\%$ over the baseline method \cite{behley2019semantickitti}. For 2D S3CNet, we conduct extra experiments on the nuScenes dataset \cite{caesar2020nuscenes}. Ground truth 2D semantic scenes are created from object bounding boxes and cropped HD map data, and aligned with LiDAR frames at an identical spatial extent and voxel resolution to SemanticKITTI. We produced 1,166,187 frames of valid LiDAR scan and semantic scene label pairs and we split the train, validation, test into a 14:3:3 ratio. The total training time of our 2D S3CNet for 50 epochs on the nuScene dataset is \(\sim\)350 hours. Table. \ref{table:2d_quantitative_result} shows that our 2D S3CNet outmatches several well-known LiDAR segmentation baselines (adapted for BEV predictions) on the segmentation component of the SSC task, with comparable results on the completion component for both the SemanticKITTI and nuScenes datasets.

\textbf{Ablation study.} We investigate the individual contribution of all components in our system. As shown in Table. \ref{table:ablation_study}, we conduct various experiments on the 2D and 3D SSC task by modifying or removing core components of the system and track the resulting effect on mean IoU and completion IoU scores on the SemanticKITTI validation set. For both the 2D and 3D networks, we observe a mean IoU drop after removing the CAM and SR modules. Since the decode blocks are mainly responsible for completion, removing SR modules results in a more severe drop in completion IoU compared to CAM which are only present in the encoder. Eliminating spatial features decreases overall performance by a large amount; the system losses geometric priors that guide completion and spatial priors that distinguish free from occluded space. Adopting Lovasz-softmax achieves the highest mean IoU increase, since it directly optimizes for the mean IoU metric (Jaccard index). In our experiments, focal loss combined with binary cross entropy loss provides no performative advantage over the baseline weighted cross entropy loss for both the 2D and 3D networks. Post-processing modules like Multi-view Fusion and Spatial Propagation Network demonstrate very high contribution to the final results - without MVF the system performance degrades to that of the focal loss baseline. A key distinction between MVF and SPN is the comparative impact of MVF on completion IoU to the SPN on segmentation mean IoU, respectively.


\begin{table}[htb]
\begin{adjustbox}{max width=\textwidth}
\begin{tabular}{|c|c|cccccccccc|c|c|}
\hline
\multirow{2}{*}{Dimension} & \multirow{2}{*}{Model}       & \multicolumn{2}{c|}{Features}                           & \multicolumn{5}{c|}{Losses}                                                                                                                                                 & \multicolumn{1}{c|}{\multirow{2}{*}{MVF}} & \multicolumn{1}{c|}{\multirow{2}{*}{SPN}} & \multicolumn{1}{c|}{\multirow{2}{*}{Data Aug.}} & \multicolumn{1}{c|}{\multirow{2}{*}{mIoU (val)}} & \multirow{2}{*}{Completion IoU (val)} \\ \cline{3-9}
                           &                              & \multicolumn{1}{c|}{Normal} & \multicolumn{1}{c|}{TSDF} & \multicolumn{1}{c|}{$\mathcal{L}_{GA}$ (Ours)} & \multicolumn{1}{c|}{WeightedCE} & \multicolumn{1}{c|}{focal} & \multicolumn{1}{c|}{Lovasz} & \multicolumn{1}{c|}{BinaryCE} & \multicolumn{1}{c|}{}                     & \multicolumn{1}{c|}{}                     & \multicolumn{1}{c|}{}                           & \multicolumn{1}{c|}{}                            &                                       \\ \hline
3D                         & \multirow{2}{*}{Full Model}  & \checkmark                  & \checkmark                & \checkmark                                     &                                 &                            &                             & \checkmark                    & \checkmark                                & \checkmark                                & \checkmark                                      & 0.3308                                            & 0.5712                                \\ \cline{1-1} \cline{3-14}
2D                         &                              & \checkmark                  & \checkmark                &                                                & \checkmark                      &                            &                             & \checkmark                    & -                                         & -                                         & \checkmark                                      & 0.2789                                           & 0.7032                                \\ \cline{1-2} \cline{3-14}
3D                         & \multirow{2}{*}{w/o CAM}      & \checkmark                  & \checkmark                & \checkmark                                     &                                 &                            &                             & \checkmark                    & \checkmark                                & \checkmark                                & \checkmark                                      & 0.2974                                           &    0.5833                             \\ \cline{1-1} \cline{3-14}
2D                         &                              & \checkmark                  & \checkmark                &                                                & \checkmark                      &                            &                             & \checkmark                    & -                                         & -                                         & \checkmark                                      & 0.2803                                           & 0.6880                                \\ \cline{1-2}\cline{3-14}
3D                         & \multirow{2}{*}{w/o SR}     & \checkmark                  & \checkmark                & \checkmark                                     &                                 &                            &                             & \checkmark                    & \checkmark                                & \checkmark                                & \checkmark                                      & 0.3005                                           & 0.5455                                \\ \cline{1-1}\cline{3-14}
2D                         &                              & \checkmark                  & \checkmark                &                                                & \checkmark                      &                            &                             & \checkmark                    & -                                         & -                                         & \checkmark                                      & 0.2604                                           & 0.6097                                \\ \cline{1-2}\cline{3-14}
3D                         & \multirow{2}{*}{w/o Feature} &                             &                           & \checkmark                                     &                                 &                            &                             & \checkmark                    & \checkmark                                & \checkmark                                & \checkmark                                      & 0.2795                                           & 0.4345                                \\ \cline{1-1}\cline{3-14}
2D                         &                              &                             &                           &                                                & \checkmark                      &                            &                             & \checkmark                    & -                                         & -                                         & \checkmark                                      & 0.2518                                           & 0.6597                                \\ \cline{1-2}\cline{3-14}
3D                         & \multirow{2}{*}{Lovasz-softmax} & \checkmark                  & \checkmark                &                                                &                                 &                            & \checkmark                  & \checkmark                    & \checkmark                                & \checkmark                                & \checkmark                                      & 0.3012                                          & 0.5934                                \\ \cline{1-1}\cline{3-14}
2D                         &                              & \checkmark                  & \checkmark                &                                                &                                 &                            & \checkmark                  & \checkmark                    & -                                         & -                                         & \checkmark                                      & 0.2633                                           & 0.6089                               \\ \cline{1-2}\cline{3-14}
3D                         & \multirow{2}{*}{Focal Loss (baseline)}  & \checkmark                  & \checkmark                &                                                &                                 & \checkmark                 &                             & \checkmark                    & \checkmark                                & \checkmark                                & \checkmark                                      & 0.2853                                           & 0.5319                                \\ \cline{1-1}\cline{3-14}
2D                         &                              & \checkmark                  & \checkmark                &                                                &                                 & \checkmark                 &                             & \checkmark                    & -                                         & -                                         & \checkmark                                      & 0.2403                                           & 0.7136                                \\ \cline{1-2}\cline{3-14}
3D                         & \multirow{2}{*}{w/o MVF}     & \checkmark                  & \checkmark                &                                                &                                 & \checkmark                 &                             & \checkmark                    &                                           & \checkmark                                & \checkmark                                      & 0.2780                                            & 0.4430                                \\ \cline{1-1}\cline{3-14}
2D                         &                              & -                           & -                         & -                                              & -                               & -                          & -                           & -                             & -                                         & -                                         & -                                               & -                                                & -                                     \\ \cline{1-2} \cline{3-14}
3D                         & \multirow{2}{*}{w/o SPN}     & \checkmark                  & \checkmark                &                                                &                                 & \checkmark                 &                             & \checkmark                    & \checkmark                                &                                             & \checkmark                                      & 0.2493                                           & 0.4902                               \\ \cline{1-1} \cline{3-14}
2D                         &                              & \checkmark                  & \checkmark                &                                                &                                 & \checkmark                 &                             & \checkmark                    & -                                         & -                                         & \checkmark                                      & 0.2303                                           & 0.6736                                \\ \hline
\end{tabular}
\end{adjustbox}
\captionsetup{font=scriptsize,labelfont=scriptsize}
\caption{Ablation study on 2D and 3D S3CNet models and core system components.}
\label{table:ablation_study}
\vspace{-10px}
\end{table}

\textbf{Data augmentation.} To increase model robustness, we integrate a series of data augmentation techniques into the training of our 2D and 3D networks. For both 2D and 3D tasks we apply uniform random cropping and dropout to the LiDAR point cloud, as well as uniform random translation of $\pm$0.1m in all three dimensions. On the 2D datasets, random rotations of $\pm 45^{\circ}$ are applied only on the yaw angle, however, we apply random rotation of $\pm10^{\circ}$ on any two of the three Euler angles (roll, pitch, yaw) at a time when training the 3D model.


\section{Conclusion}
\label{sec:conclusion}


In this paper, we presented a Sparse Semantic Scene Completion Network, S3CNet, capable of efficiently reconstructing large outdoor scenes and predicting semantic voxel-wise labels from a single LiDAR scan. To complement S3CNet, we designed a novel geometric-aware sparse tensor segmentation loss that promotes class-consistent predictions in homogeneous regions while encouraging locally structured multi-class segmentations along object boundaries. In combination with Multi-View Fusion and Spatial Propagation post-processing modules, our method achieves state-of-the-art results on the SemanticKITTI test set benchmark by a large margin. We also adapt several leading LiDAR segmentation networks as baselines for 2D semantic scene completion, and demonstrate that the 2D variant of S3CNet outperforms these baselines on two large-scale datasets. Furthermore, we include a detailed ablation study highlighting the contribution of each individual component to the overall system. Future work holds the extension of our sparse tensor method to real-time speeds with improved performance across the board, additional investigation on useful spatial feature encodings, and a learning-based multi-view fusion technique to enable end-to-end learning.



\clearpage
\acknowledgments{We would like to thank Professor John K. Tsotsos for reviewing this work, as well as the anonymous reviewers for their valuable suggestions.}


\bibliography{example}  

\section{Appendix A: Extra Qualitative Results}



We present additional qualitative results on the SemanticKITTI training set. As we can see in Fig. \ref{fig:qualitative_result}, our model well captures both completion and semantic segmentation characteristics of different scenes. Because a single ground truth label was constructed from LiDAR scans across several time steps (in order to densify the scene), dynamic objects were filtered out to avoid labelling noise. For instance, in the bottom-right most sample, while the bus was filtered out of the ground-truth scene, our well engineered features enabled our model to detect the object with the correct class label. Another example of this is visible in the top-right most sample, where our model detects a moving bicyclist in front of a vehicle. These object labels are learned from static scenes, but are successfully inferred in dynamic scenes which reflects positively on our model's generalization ability.

\begin{figure}[tbh]
    \centering
    \includegraphics[width=\linewidth]{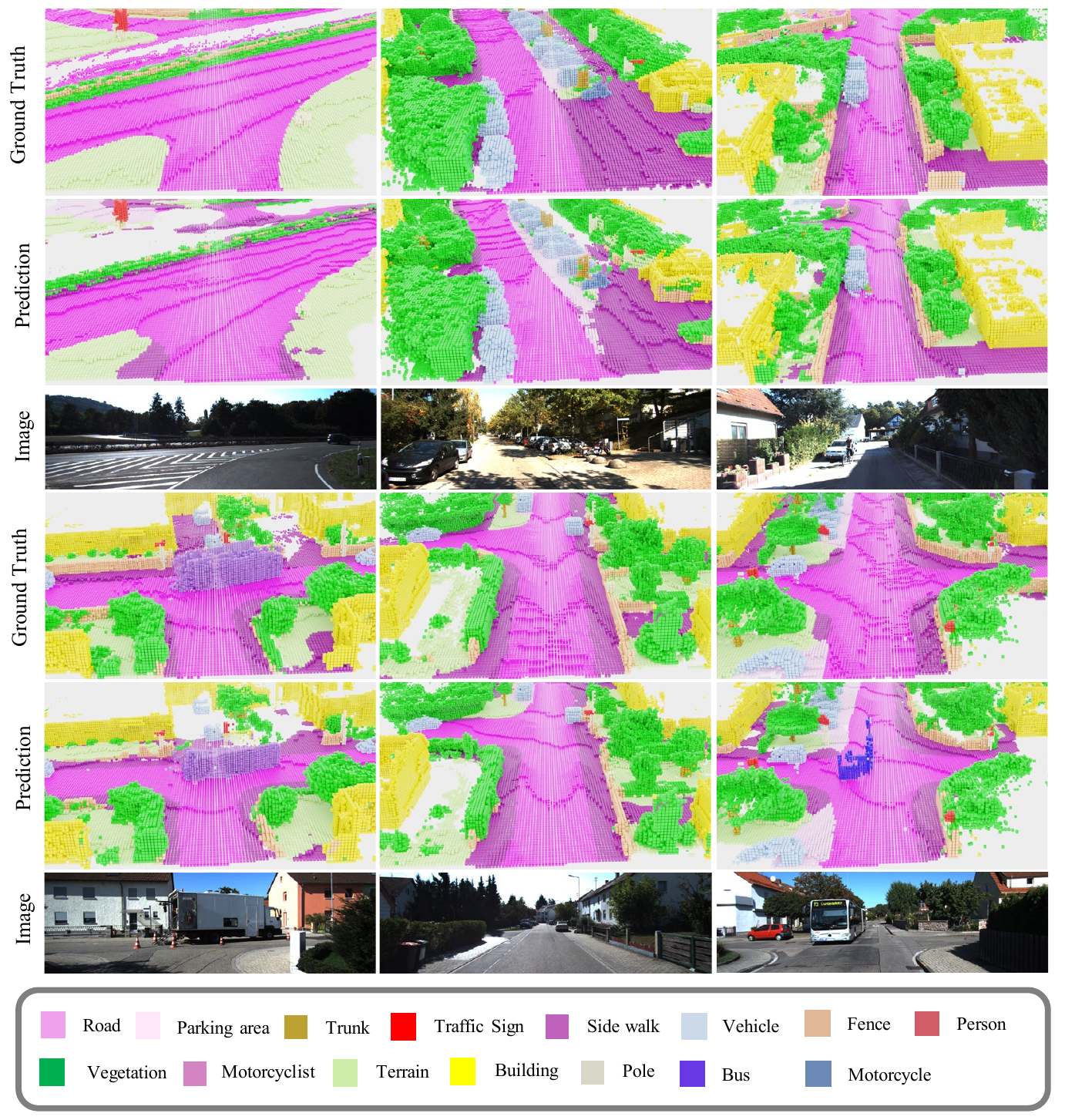}
    \caption{Qualitative results of S3CNet in SemanticKITTI dataset.}
    \label{fig:qualitative_result}
\end{figure}

\begin{figure}[htb]
    \centering
    \includegraphics[width=\linewidth]{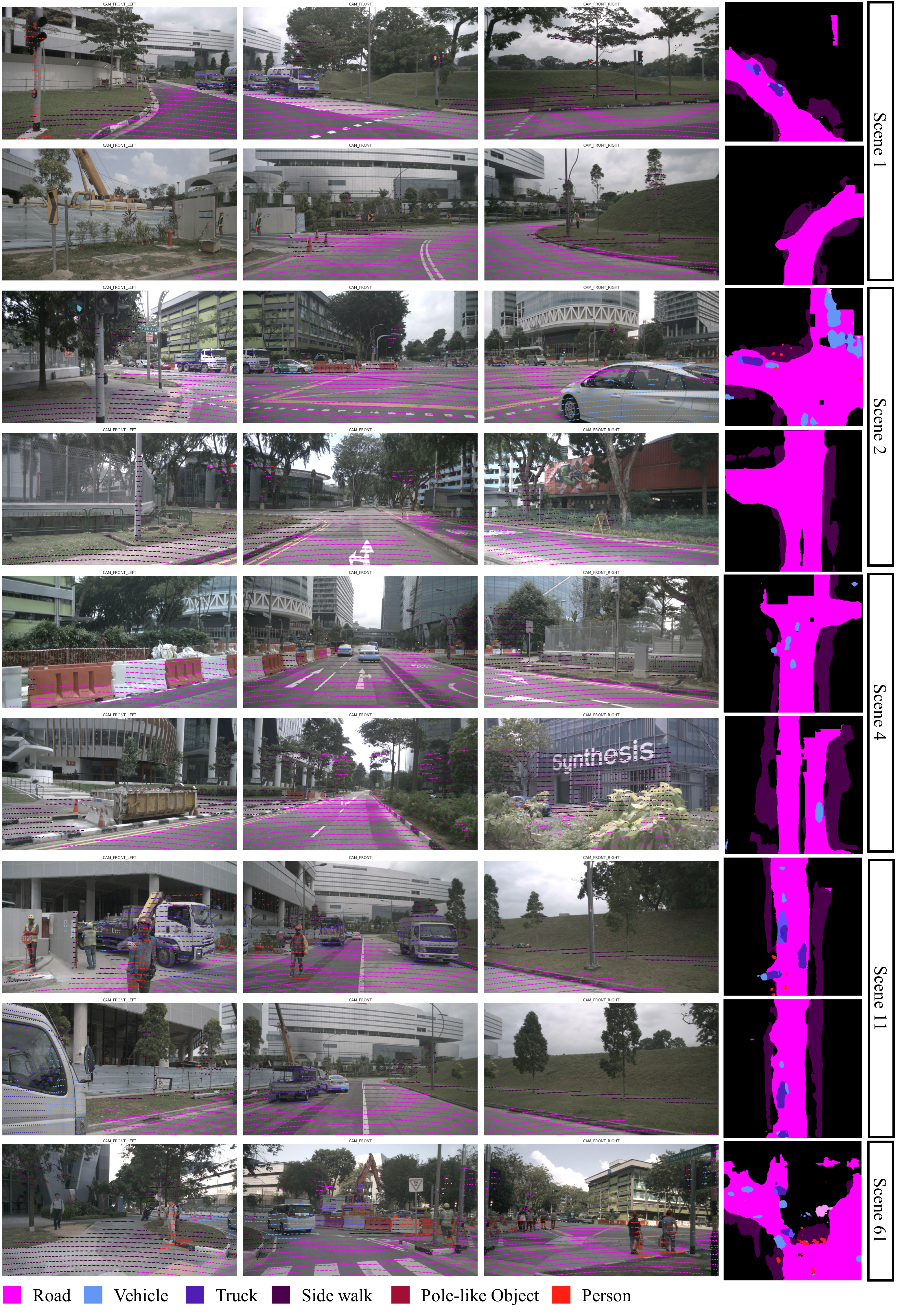}
    \caption{2D (BEV map) Semantic scene completion qualitative results in NuScenes dataset. }
    \label{fig:2d_bev_nuscenes}
\end{figure}

NuScenes 2D qualitative results: we project the 2D semantic BEV map back to 3D lidar points according the respective x and y coordinates, and overlay the semantic point cloud data on the rgb images to show the model's qualitative results. As we can see from Fig. \ref{fig:2d_bev_nuscenes}, our predictions cover most of the road surface and precisely detects vehicles, pole-like objects and pedestrians. 

\section{Appendix B: Experiment Configurations}

We provide the training scheme for all 2D experiments in Table. \ref{table: 2d_training_configs}, which includes the adapted LiDAR segmentation baselines that have undergone multiple optimization iterations to achieve the stated results on the SemanticKITTI and NuScenes datasets. Note that all experiments use identical input data and augmentation configurations; changes only occur to the model, the loss function, optimizer, scheduler, and supporting hyperparameters - much of which are based on the training scheme proposed by the original works.

\begin{table}[htb]
\begin{adjustbox}{max width=\textwidth}
\begin{tabular}{ccccc}
\hline
\multicolumn{1}{|c|}{Model}                       & \multicolumn{1}{c|}{Loss Function}                                 & \multicolumn{1}{c|}{Optimizer}                & \multicolumn{1}{c|}{Scheduler}             & \multicolumn{1}{c|}{Rates}          \\ \hline
\multicolumn{1}{|c|}{\multirow{4}{*}{HD-UNet \cite{yang2018hdnet}}}    & \multicolumn{1}{c|}{\multirow{4}{*}{Weighted CE}}                  & \multicolumn{1}{c|}{\multirow{4}{*}{SGD}}     & \multicolumn{1}{c|}{\multirow{4}{*}{Step}} & \multicolumn{1}{c|}{lr = 0.01-0.02} \\ \cline{5-5} 
\multicolumn{1}{|c|}{}                            & \multicolumn{1}{c|}{}                                              & \multicolumn{1}{c|}{}                         & \multicolumn{1}{c|}{}                      & \multicolumn{1}{c|}{dr = 0.1}       \\ \cline{5-5} 
\multicolumn{1}{|c|}{}                            & \multicolumn{1}{c|}{}                                              & \multicolumn{1}{c|}{}                         & \multicolumn{1}{c|}{}                      & \multicolumn{1}{c|}{m=0.9}          \\ \cline{5-5} 
\multicolumn{1}{|c|}{}                            & \multicolumn{1}{c|}{}                                              & \multicolumn{1}{c|}{}                         & \multicolumn{1}{c|}{}                      & \multicolumn{1}{c|}{wd=0}           \\ \hline
\multicolumn{1}{|c|}{\multirow{4}{*}{PointSeg \cite{wang2018pointseg}}}   & \multicolumn{1}{c|}{\multirow{4}{*}{Weighted CE}}                  & \multicolumn{1}{c|}{\multirow{4}{*}{Adagrad}} & \multicolumn{1}{c|}{\multirow{4}{*}{Step}} & \multicolumn{1}{c|}{lr = 0.005}     \\ \cline{5-5} 
\multicolumn{1}{|c|}{}                            & \multicolumn{1}{c|}{}                                              & \multicolumn{1}{c|}{}                         & \multicolumn{1}{c|}{}                      & \multicolumn{1}{c|}{dr = 0.1}       \\ \cline{5-5} 
\multicolumn{1}{|c|}{}                            & \multicolumn{1}{c|}{}                                              & \multicolumn{1}{c|}{}                         & \multicolumn{1}{c|}{}                      & \multicolumn{1}{c|}{m=0.9}          \\ \cline{5-5} 
\multicolumn{1}{|c|}{}                            & \multicolumn{1}{c|}{}                                              & \multicolumn{1}{c|}{}                         & \multicolumn{1}{c|}{}                      & \multicolumn{1}{c|}{wd=0.0005}      \\ \hline
\multicolumn{1}{|c|}{\multirow{4}{*}{SSv2 \cite{wu2019squeezesegv2}}}       & \multicolumn{1}{c|}{\multirow{4}{*}{Focal Loss}}                   & \multicolumn{1}{c|}{\multirow{4}{*}{SGD}}     & \multicolumn{1}{c|}{\multirow{4}{*}{Step}} & \multicolumn{1}{c|}{lr = 0.01}      \\ \cline{5-5} 
\multicolumn{1}{|c|}{}                            & \multicolumn{1}{c|}{}                                              & \multicolumn{1}{c|}{}                         & \multicolumn{1}{c|}{}                      & \multicolumn{1}{c|}{dr = 0.1}       \\ \cline{5-5} 
\multicolumn{1}{|c|}{}                            & \multicolumn{1}{c|}{}                                              & \multicolumn{1}{c|}{}                         & \multicolumn{1}{c|}{}                      & \multicolumn{1}{c|}{m=0.9}          \\ \cline{5-5} 
\multicolumn{1}{|c|}{}                            & \multicolumn{1}{c|}{}                                              & \multicolumn{1}{c|}{}                         & \multicolumn{1}{c|}{}                      & \multicolumn{1}{c|}{wd=0.0001}      \\ \hline
\multicolumn{1}{|c|}{\multirow{2}{*}{DBLiDARNet \cite{dewan2019deeptemporalseg}}} & \multicolumn{1}{c|}{\multirow{2}{*}{Weighted CE}}                  & \multicolumn{1}{c|}{\multirow{2}{*}{Adam}}    & \multicolumn{1}{c|}{\multirow{2}{*}{None}} & \multicolumn{1}{c|}{lr = 0.0001}    \\ \cline{5-5} 
\multicolumn{1}{|c|}{}                            & \multicolumn{1}{c|}{}                                              & \multicolumn{1}{c|}{}                         & \multicolumn{1}{c|}{}                      & \multicolumn{1}{c|}{wd=0.0005}      \\ \hline
\multicolumn{1}{|c|}{\multirow{4}{*}{SalsaNext \cite{cortinhal2020salsanext}}}  & \multicolumn{1}{c|}{\multirow{4}{*}{Weighted CE + Lovasz Softmax}} & \multicolumn{1}{c|}{\multirow{4}{*}{SGD}}     & \multicolumn{1}{c|}{\multirow{4}{*}{Exp}}  & \multicolumn{1}{c|}{lr=0.05}        \\ \cline{5-5} 
\multicolumn{1}{|c|}{}                            & \multicolumn{1}{c|}{}                                              & \multicolumn{1}{c|}{}                         & \multicolumn{1}{c|}{}                      & \multicolumn{1}{c|}{dr=0.01}        \\ \cline{5-5} 
\multicolumn{1}{|c|}{}                            & \multicolumn{1}{c|}{}                                              & \multicolumn{1}{c|}{}                         & \multicolumn{1}{c|}{}                      & \multicolumn{1}{c|}{m=0.9}          \\ \cline{5-5} 
\multicolumn{1}{|c|}{}                            & \multicolumn{1}{c|}{}                                              & \multicolumn{1}{c|}{}                         & \multicolumn{1}{c|}{}                      & \multicolumn{1}{c|}{wd=0.0001}      \\ \hline
\multicolumn{1}{|c|}{\multirow{4}{*}{Ours}}       & \multicolumn{1}{c|}{\multirow{4}{*}{Focal Loss + Weighted CE + BCE}}                             & \multicolumn{1}{c|}{\multirow{4}{*}{SGD}}     & \multicolumn{1}{c|}{\multirow{4}{*}{Exp}}  & \multicolumn{1}{c|}{lr=0.001}       \\ \cline{5-5} 
\multicolumn{1}{|c|}{}                            & \multicolumn{1}{c|}{}                                              & \multicolumn{1}{c|}{}                         & \multicolumn{1}{c|}{}                      & \multicolumn{1}{c|}{dr=0.01}        \\ \cline{5-5} 
\multicolumn{1}{|c|}{}                            & \multicolumn{1}{c|}{}                                              & \multicolumn{1}{c|}{}                         & \multicolumn{1}{c|}{}                      & \multicolumn{1}{c|}{m=0.9}          \\ \cline{5-5} 
\multicolumn{1}{|c|}{}                            & \multicolumn{1}{c|}{}                                              & \multicolumn{1}{c|}{}                         & \multicolumn{1}{c|}{}                      & \multicolumn{1}{c|}{wd=0.0005}      \\ \hline
\multicolumn{5}{l}{lr: learning rate}                                                                                                                                                                                                                     \\
\multicolumn{5}{l}{dr: decay rate}                                                                                                                                                                                                                        \\
\multicolumn{5}{l}{m: momentum}                                                                                                                                                                                                                           \\
\multicolumn{5}{l}{wd: weight decay}                                                                                                                             
\end{tabular}
\end{adjustbox}
\caption{Training configurations for 2D Semantic Scene Completion models on SemanticKITTI and NuScenes datasets.}
\label{table: 2d_training_configs}
\end{table}

We also list our experiment configurations including the loss function, hyperparameters, scheduler, and provide the best validation mean IoU for all 3D model experiments (see Table. \ref{table: 3d_training_configs}). We implemented a list of state-of-the-art methods (not currently on the SemanticKITTI benchmark) \cite{song2017semantic}\cite{li2020anisotropic}\cite{li2019rgbd}\cite{li2019depth} and trained them on the SemanticKITTI dataset for 50 epochs. Amongst the competitors, our model achieved the best mean IoU on the validation split (sequence 08).

\begin{table}[tbh]
\begin{adjustbox}{max width=\textwidth}
\begin{tabular}{cccccl}
\hline
\multicolumn{1}{|c|}{Model}                   & \multicolumn{1}{c|}{Loss Function}                & \multicolumn{1}{c|}{Optimizer}            & \multicolumn{1}{c|}{Scheduler}             & \multicolumn{1}{c|}{Rates}      & \multicolumn{1}{l|}{Best mIoU}                   \\ \hline
\multicolumn{1}{|c|}{\multirow{4}{*}{SSCNet \cite{song2017semantic}}} & \multicolumn{1}{c|}{\multirow{4}{*}{Weighted CE}} & \multicolumn{1}{c|}{\multirow{4}{*}{SGD}} & \multicolumn{1}{c|}{\multirow{4}{*}{Step}} & \multicolumn{1}{c|}{lr = 0.01}  & \multicolumn{1}{l|}{\multirow{4}{*}{0.182}} \\ \cline{5-5}
\multicolumn{1}{|c|}{}                        & \multicolumn{1}{c|}{}                             & \multicolumn{1}{c|}{}                     & \multicolumn{1}{c|}{}                      & \multicolumn{1}{c|}{dr = 0.1}   & \multicolumn{1}{l|}{}                       \\ \cline{5-5}
\multicolumn{1}{|c|}{}                        & \multicolumn{1}{c|}{}                             & \multicolumn{1}{c|}{}                     & \multicolumn{1}{c|}{}                      & \multicolumn{1}{l|}{m=0.9}      & \multicolumn{1}{l|}{}                       \\ \cline{5-5}
\multicolumn{1}{|c|}{}                        & \multicolumn{1}{c|}{}                             & \multicolumn{1}{c|}{}                     & \multicolumn{1}{c|}{}                      & \multicolumn{1}{l|}{wd=0.0005}  & \multicolumn{1}{l|}{}                       \\ \hline
\multicolumn{1}{|c|}{\multirow{4}{*}{AICNet \cite{li2020anisotropic}}} & \multicolumn{1}{c|}{\multirow{4}{*}{Weighted CE}} & \multicolumn{1}{c|}{\multirow{4}{*}{SGD}} & \multicolumn{1}{c|}{\multirow{4}{*}{Step}} & \multicolumn{1}{c|}{lr = 0.015} & \multicolumn{1}{l|}{\multirow{4}{*}{0.215}} \\ \cline{5-5}
\multicolumn{1}{|c|}{}                        & \multicolumn{1}{c|}{}                             & \multicolumn{1}{c|}{}                     & \multicolumn{1}{c|}{}                      & \multicolumn{1}{c|}{dr = 0.1}   & \multicolumn{1}{l|}{}                       \\ \cline{5-5}
\multicolumn{1}{|c|}{}                        & \multicolumn{1}{c|}{}                             & \multicolumn{1}{c|}{}                     & \multicolumn{1}{c|}{}                      & \multicolumn{1}{l|}{m=0.9}      & \multicolumn{1}{l|}{}                       \\ \cline{5-5}
\multicolumn{1}{|c|}{}                        & \multicolumn{1}{c|}{}                             & \multicolumn{1}{c|}{}                     & \multicolumn{1}{c|}{}                      & \multicolumn{1}{l|}{wd=0.0005}  & \multicolumn{1}{l|}{}                       \\ \hline
\multicolumn{1}{|c|}{\multirow{4}{*}{DDRNet \cite{li2019rgbd}}} & \multicolumn{1}{c|}{\multirow{4}{*}{Weighted CE}} & \multicolumn{1}{c|}{\multirow{4}{*}{SGD}} & \multicolumn{1}{c|}{\multirow{4}{*}{Exp}}  & \multicolumn{1}{c|}{lr = 0.02}  & \multicolumn{1}{l|}{\multirow{4}{*}{0.193}} \\ \cline{5-5}
\multicolumn{1}{|c|}{}                        & \multicolumn{1}{c|}{}                             & \multicolumn{1}{c|}{}                     & \multicolumn{1}{c|}{}                      & \multicolumn{1}{c|}{dr = 0.1}   & \multicolumn{1}{l|}{}                       \\ \cline{5-5}
\multicolumn{1}{|c|}{}                        & \multicolumn{1}{c|}{}                             & \multicolumn{1}{c|}{}                     & \multicolumn{1}{c|}{}                      & \multicolumn{1}{l|}{m=0.9}      & \multicolumn{1}{l|}{}                       \\ \cline{5-5}
\multicolumn{1}{|c|}{}                        & \multicolumn{1}{c|}{}                             & \multicolumn{1}{c|}{}                     & \multicolumn{1}{c|}{}                      & \multicolumn{1}{l|}{wd=0.0001}  & \multicolumn{1}{l|}{}                       \\ \hline
\multicolumn{1}{|c|}{\multirow{4}{*}{PALNet \cite{li2019depth}}} & \multicolumn{1}{c|}{\multirow{4}{*}{Position Importance Aware Loss}} & \multicolumn{1}{c|}{\multirow{4}{*}{SGD}} & \multicolumn{1}{c|}{\multirow{4}{*}{Exp}}  & \multicolumn{1}{c|}{lr = 0.02}  & \multicolumn{1}{l|}{\multirow{4}{*}{0.255}} \\ \cline{5-5}
\multicolumn{1}{|c|}{}                        & \multicolumn{1}{c|}{}                             & \multicolumn{1}{c|}{}                     & \multicolumn{1}{c|}{}                      & \multicolumn{1}{c|}{dr = 0.05}  & \multicolumn{1}{l|}{}                       \\ \cline{5-5}
\multicolumn{1}{|c|}{}                        & \multicolumn{1}{c|}{}                             & \multicolumn{1}{c|}{}                     & \multicolumn{1}{c|}{}                      & \multicolumn{1}{l|}{m=0.9}      & \multicolumn{1}{l|}{}                       \\ \cline{5-5}
\multicolumn{1}{|c|}{}                        & \multicolumn{1}{c|}{}                             & \multicolumn{1}{c|}{}                     & \multicolumn{1}{c|}{}                      & \multicolumn{1}{l|}{wd=0.0001}  & \multicolumn{1}{l|}{}                       \\ \hline
\multicolumn{1}{|c|}{\multirow{4}{*}{Ours}} & \multicolumn{1}{c|}{\multirow{4}{*}{Geo-Aware Loss + Binary Cross Entropy}} & \multicolumn{1}{c|}{\multirow{4}{*}{SGD}} & \multicolumn{1}{c|}{\multirow{4}{*}{Exp}}  & \multicolumn{1}{c|}{lr = 0.025}  & \multicolumn{1}{l|}{\multirow{4}{*}{0.303}} \\ \cline{5-5}
\multicolumn{1}{|c|}{}                        & \multicolumn{1}{c|}{}                             & \multicolumn{1}{c|}{}                     & \multicolumn{1}{c|}{}                      & \multicolumn{1}{c|}{dr = 0.1}   & \multicolumn{1}{l|}{}                       \\ \cline{5-5}
\multicolumn{1}{|c|}{}                        & \multicolumn{1}{c|}{}                             & \multicolumn{1}{c|}{}                     & \multicolumn{1}{c|}{}                      & \multicolumn{1}{l|}{m=0.9}      & \multicolumn{1}{l|}{}                       \\ \cline{5-5}
\multicolumn{1}{|c|}{}                        & \multicolumn{1}{c|}{}                             & \multicolumn{1}{c|}{}                     & \multicolumn{1}{c|}{}                      & \multicolumn{1}{l|}{wd=0.0001}  & \multicolumn{1}{l|}{}                       \\ \hline
\multicolumn{6}{l}{lr: learning rate}                                                                                                                                                                                                                                      \\
\multicolumn{6}{l}{dr: decay rate}                                                                                                                                                                                                                                         \\
\multicolumn{6}{l}{m: momentum}                                                                                                                                                                                                                                            \\
\multicolumn{6}{l}{wd: weight decay}                                                                                                                                                                                                                                      
\end{tabular}
\end{adjustbox}
\caption{Training configurations for 3D Semantic Scene Completion models in SemanticKITTI dataset.}
\label{table: 3d_training_configs}
\end{table}

\end{document}